\documentclass[Afour,sageh,times]{sagej}

\usepackage{moreverb,url}

\usepackage[colorlinks,bookmarksopen,bookmarksnumbered,citecolor=red,urlcolor=red]{hyperref}

\usepackage{graphics} 
\usepackage{caption}
\usepackage{epsfig} 
\usepackage{amsmath} 
\usepackage{amssymb}  
\usepackage{mathtools}
\usepackage{bm}
\usepackage{algorithm,algcompatible,algpseudocode} 
\usepackage[dvipsnames,table,svgnames]{xcolor}

\usepackage{multicol}
\usepackage{multirow}
\usepackage{booktabs} 

\usepackage{cleveref}

\usepackage[toc]{appendix}

\def\volumeyear{2024}

\begin{document}

\runninghead{Lin, et al.}

\title{Navigation and 3D Surface Reconstruction from Passive\\
Whisker Sensing}

\author{Michael A. Lin\affilnum{1}, Hao Li\affilnum{1}, Chengyi Xing\affilnum{1} and Mark R. Cutkosky\affilnum{1}}

\affiliation{\affilnum{1}Stanford University, CA, USA}

\corrauth{Michael A. Lin, Biomimetic and Dexterous Manipulation Lab,
Stanford University,
Stanford, California,
USA}

\email{[mlinyang,li2053,chengyix,cutkosky]@stanford.edu}


\begin{abstract}
Whiskers provide a way to sense surfaces in the immediate environment without disturbing it. In this paper we present a method for using highly flexible, curved, passive whiskers mounted along a robot arm to gather sensory data as they brush past objects during normal robot motion. The information is useful both for guiding the robot in cluttered spaces and for reconstructing the exposed faces of objects. Surface reconstruction depends on accurate localization of contact points along each whisker. We present an algorithm based on Bayesian filtering that rapidly converges to within 1\,mm of the actual contact locations. The piecewise-continuous history of contact locations from each whisker allows for accurate reconstruction of curves on object surfaces. Employing multiple whiskers and traces, we are able to produce an occupancy map of proximal objects. 
\end{abstract}

\keywords{Perception for Grasping and Manipulation, Soft Sensors and Actuators, Force and Tactile Sensing, Biologically-Inspired Robots, Collision Avoidance}

\maketitle

\section{Introduction}

Whiskers are effective for operating in confined and cluttered environments---especially when vision is poor---and can provide valuable information about objects and surfaces as they contact and brush past them. 
Not surprisingly, they are used widely across the animal kingdom \cite{prescott2016vibrissal,boublil2021mechanosensory}. In some cases (e.g. rats) animals employ sophisticated active whisking and a specialized sensory cortex for processing whisker stimulii ~\cite{carvell1990biometric}. 
In other cases, such as the vibrissae on the lower limbs of cats, they are mostly passive and provide information for gracefully negotiating complex three-dimensional environments. It has been shown also that the intrinsic curvature of whiskers plays a pivotal role in influencing the mechanical signals they transmit \cite{luo2023intrinsic}.

Although they are less widely used than other sensing modalities, whiskers have also been demonstrated in robotics. Relevant prior work is summarized in the next section, but first we consider some general characteristics of whisker sensing.
Like tactile sensors, they provide information about contacts and surfaces at close range and when vision is occluded. Unlike most tactile sensors, they can contact even very light objects without disturbing them. This ability to sense the environment without changing its state is useful for state estimation as the state of the environment remains unchanged. In this regard, whiskers can also be compared to non-contact proximity sensing, for example using ultrasound, or capacitive, or optical transducers \cite{navarro2021proximity}. Compared to many non-contact sensors, whiskers have a small receptive field (i.e., they typically sense a single point of contact at any time), which may be useful to isolate discontinuous object features. In addition, although they do not affect the environment, they are affected by it. Changes in object texture, for example, can affect whisker signals on the high frequency ranges. Depending on the whisker curvature, they can also react differently than a non-contact proximity sensor when passing off the corner of one object and onto another. An advantage we show experimentally, is that whiskers are more robust to object surface reflectivity and translucency compared to many of the existing non-contact proximity sensors.

\begin{figure}[thpb!]
\vspace{4 pt}
  \centering
  \includegraphics[width=\linewidth]{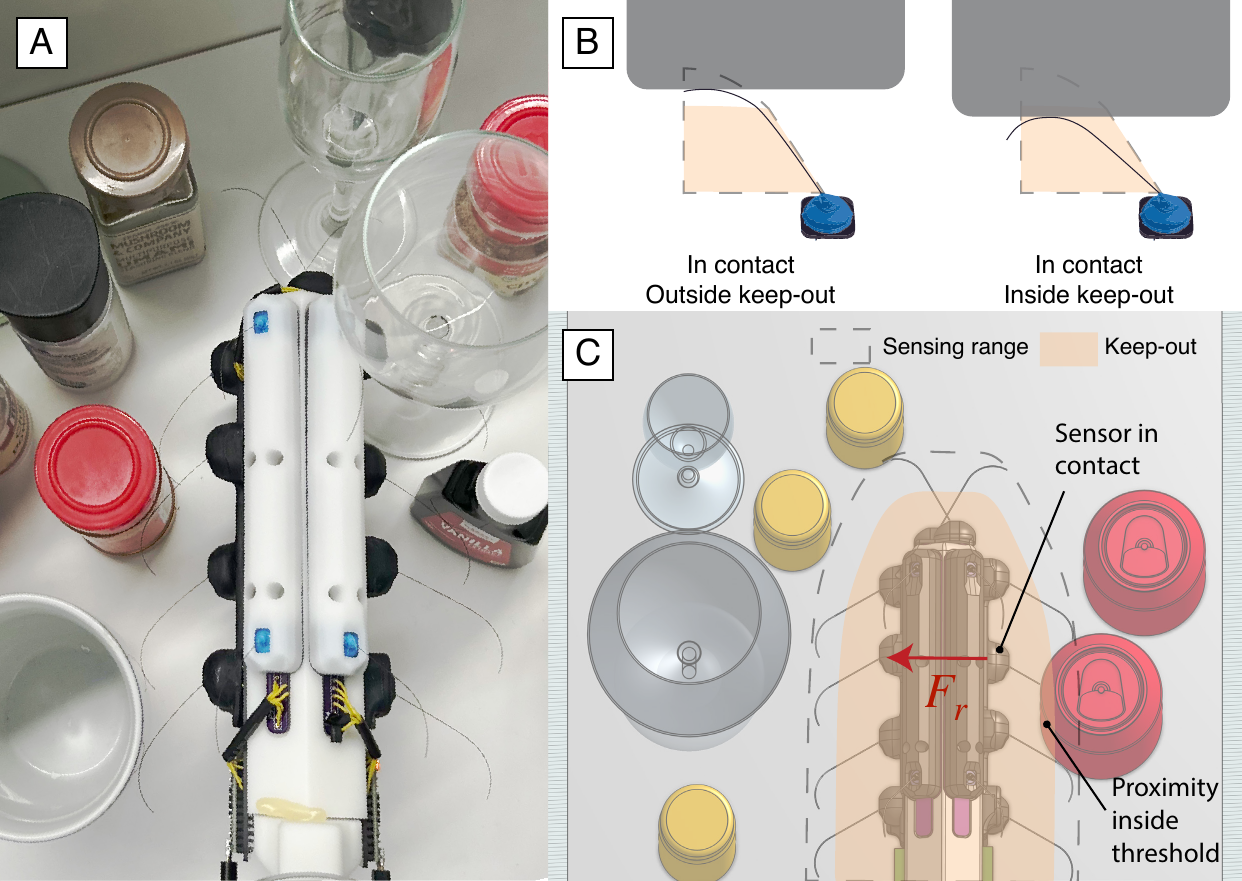}
  \caption{A)Instruments robot end-effector with an array of 16 semi-curved whisker sensors. B)Illustrates the sensing region of a semi-curved sensor. Shaded region approximately defines a threshold or keep-out region based on the amount of deflection measured on the sensor. C)Shows the combined sensing region and keep-out region of a sensorized end-effector. Contacts within the threshold generate a repelling force $F_r$.}
  \label{fig:system}
 \vspace{-2em}s
\end{figure}

The motivating application for the whiskers and perception method reported here is robots that operate in cluttered environments.  
Reaching into a cupboard or a refrigerator full of objects are examples of tasks where contacts may happen frequently and even unexpectedly, both at the end-effector and along the forearm (\Cref{fig:system}). Sensing the locations and forces of contacts has been recognized as helpful for perception \cite{jain2013reaching,wang2020contact,bohg2010strategies,petrovskaya2011global} enabling access to locations that may be occluded from visual sensors. However, sensing through contact when interacting with free-standing objects that are small and light (e.g. a nearly empty bottle of pills or spices) is challenging because the action of making contact will likely change the object's state. This issue is pronounced when contacts occur on parts of a robot with high mechanical impedance. Enabling robots to sense objects unobtrusively allows for the assumption that the environment remains static through interactions. Consequently, the history of contact measurements becomes coherent and integrable. To achieve unobtrusive tactile sensing, prior work has shown that minimizing the robot's inertial properties at the end-effector is useful \cite{bhatia2019direct,wang2020contact,lin2021exploratory}. However, a different approach is needed to detect and minimize disturbances arising from contacts along the arm.

In our approach, we use soft, curved whiskers mounted along a robot arm and we use the robot's accurate proprioception in combination with sensor measurements to localize contacts along the whisker and gather information about the environment.

\begin{figure}[htpb]
  \centering
  \includegraphics[width=0.8\linewidth]{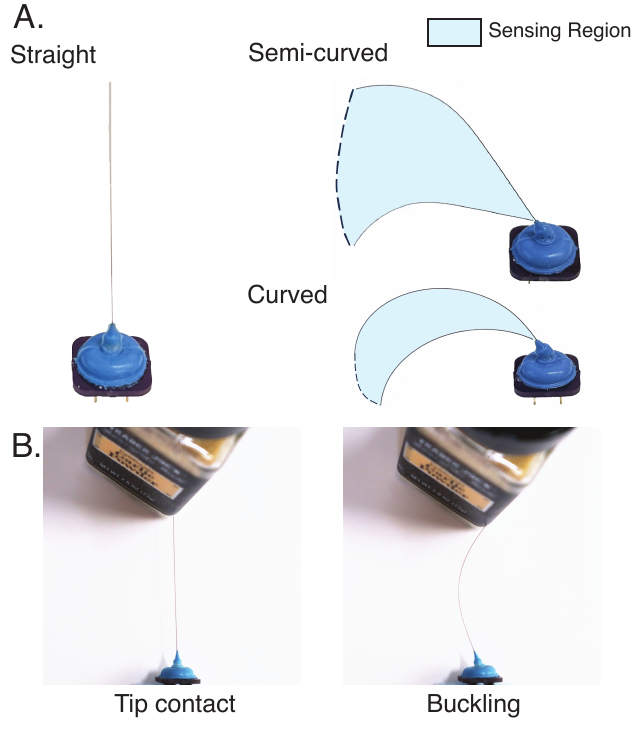}
  \caption{
 Curved whiskers with different geometries and sensing regions (A). Depending on motion direction, a straight whisker may buckle, making interpretation more difficult (B).}
    \vspace{-1em}
  \label{fig:geometry}
\end{figure}

A number of challenges arise in 
passive whisker use.
First is that whisker motion and deflection are subject to arm motion, which is typically intended to control an end-effector rather than to provide exploratory sensing. Interaction with objects will often be limited to one sustained contact as opposed to multiple probing actions. The state estimation method should be able to process this arbitrary motion and whisker deflection to infer contact locations quickly. A second challenge is that passive interactions with the world can cause a whisker to be deflected in any direction. Therefore, contact localization has to be computed in 3D. A third challenge is that as a whisker is moved in arbitrary directions a straight whisker may catch its tip on object surfaces and buckle (\Cref{fig:geometry}), making the sensor signal difficult to interpret. A curved whisker can avoid this buckling effect, but contact localization becomes more challenging. 

\textbf{Contributions}: We present methods for employing arrays of flexible whiskers for navigation and partial surface reconstruction. In this work we use pre-curved whiskers of super-elastic nitinol mounted along the arm of a robot; fabrication methods are presented in an earlier paper \cite{lin2022whisker}. To evaluate the effectiveness of these whiskers to perceive nearby objects and surfaces, we conducted preliminary experiments comparing their performance to other proximity sensors. In comparative tests, we report the accuracies of distance measurements to surfaces using these whiskers and with optical and ultrasonic non-contact sensors. These tests are relevant for obstacle avoidance and navigation. We further develop an algorithm that can precisely determine 3D contact location on the whisker in real-time. The main components of this algorithm include: 1. a calibration-based sensor model using a Gaussian Process Regression (GPR) method with a Thin Plate kernel and demonstrate the advantage of this kernel over other popular GPR kernels, and 2. a Bayesian filter that combines the sensor and proprioceptive data to infer contact location in real-time. We implement three different Bayesian filters (Extended Kalman Filter, Unscented Kalman Filter and Particle Filter), showing that these methods can perform better than a baseline method \cite{solomon2010extracting}, with the UKF being the best performing filter in tests. Using this approach, we demonstrate the ability to combine robot proprioception and sensor measurements to accurately track contact locations over time and to partially reconstruct object surfaces using passive whisker contact. In addition to contact localization, we also present a method to integrate sequences of contact location estimates using Bayesian Hilbert Maps. We show how this occupancy map method can serve multiple purpose including: 1. combine sensory data from multiple sensors, 2. provide a more accurate prior distribution to initialize a Bayesian Filtering run upon first contact, and 3. integrate sensory data over time and interpolate the surface structure of surrounding space.

Some highlighted differences of this work compared to a preceding conference paper \cite{lin2022whisker} are that we develop algorithms to track contact locations in 3D, not constraining them to the primary curvature plane of the sensor. We also develop methods for modeling the sensor with Gaussian Process Regression that is more robust to tracking divergence. Finally, we present results on integrating histories of contacts to estimate object surfaces.

\section{Related work}
Whiskers have been employed to perceive diverse types of environmental information, including contact, force, and air currents or flows \cite{lin2022whisker, zhang2023novel, liu2023artificial}. This innate capacity for spatial and tactile awareness has inspired numerous applications in robotics, particularly in navigating and understanding unstructured environments \cite{suresh2021tactile, lin2022whisker}. 

\label{sec:related}
\subsection{Contact interpretation}
The sensing method presented here builds upon prior work on perception of unstructured environments through contacts \cite{petrovskaya2011global,bohg2010strategies,wang2020contact,koval2015pose,manuelli2016localizing,suresh2021tactile,lin2021exploratory}. A common challenge when sensing free-standing objects is that the act of contact sensing often will change the state of the object. For objects of known shapes, Koval \emph{et al.} developed a Particle Filtering approach to estimate object location through a sequence of pushes to collapse the belief distribution \cite{koval2015pose}. Suresh \emph{et al.} expanded on this work by posing the problem as a Simultaneous Localization And Mapping (SLAM) problem and were able to estimate both shape and location of objects \cite{suresh2021tactile}. While these methods work when interacting with isolated objects, the pushing approach is more challenging when an object is amidst clutter which constrains both the object and robot arm.

\subsection{Unobtrusive contact sensing}
\label{sec:related-unobtrusive}
Sensing through non-intrusive contacts has the advantage of objects remaining static, making state estimation easier. The most common of such perception methods is vision, however, RGB-D cameras are not well suited for close range sensing as is typically necessary when reaching into confined spaces with objects. Some work has investigated using close range proximity sensing \cite{hsiao2009proximity,tsuji2019proximity,schlegl2013virtual}, but these methods do not perform well when sensing specular or transparent surfaces for optical transducers, or may be susceptible to variations in materials properties for magnetic and capacitive transducers. Sensing through mechanical contact is not affected by these problems.

\subsection{Whiskers for perception}
 Some investigations have addressed whisker- or antenna-based sensing in robotics. Early work by Kaneko \emph{et al.} showed a method of active probing where a flexible antenna is rotated by actuators at one end to make contact with objects while estimating contact location with measured rotational compliance \cite{kaneko1998active}. Subsequent efforts have addressed improved contact localization for objects of varying shapes \cite{solomon2010extracting,kim2007biomimetic,merker2021vibrissa,gomez2024bioinspired,gomez2023robominer}. Using 3-DOF force/torque sensing at the base (two bending torques and one axial force) it is possible to deduce contact locations from single measurements \cite{caliwhisker1,caliwhisker2, huet2017tactile,emnett2018novel,nguyen2020contact}. 
However, these methods require either solving a nonlinear system of differential equations or providing a unique mapping of transducer signals to contact locations, leading to computational complexity and limited resolution under realistic conditions when an arm is moving among small and light objects.

\subsection{Safe collision-avoidance navigation}
For the task of reaching in an unstructured space, an advantage of using non-obtrusive sensing is that it can be directly used for collision-free navigation. Typically, collision avoidance navigation in clutter is done by sensing through rigid or soft collisions \cite{saund2019blindfolded, saloutos2023design, gruebele2020stretchable,gruebele2021stretchable}, model-based control \cite{jain2013reaching} and path planning strategies \cite{muhayyuddin2018randomized}. However, relying on forceful collisions for navigation can be challenging as the safety of the solution depends on speed of robot motion, sensing, control latency, and compliance of the system. Using optical or acoustic proximity sensing has robustness limitations to object translucency and texture as mentioned in \Cref{sec:related-unobtrusive}. Using highly compliant structures to sense surrounding spaces, on the other hand, overcomes these limitations while effectively still achieving a safe interaction with surrounding objects.


\section{Whisker sensing principle}
\begin{figure}[bh]
    \vspace{-1em}
  \centering
  \includegraphics[width=0.7\linewidth]{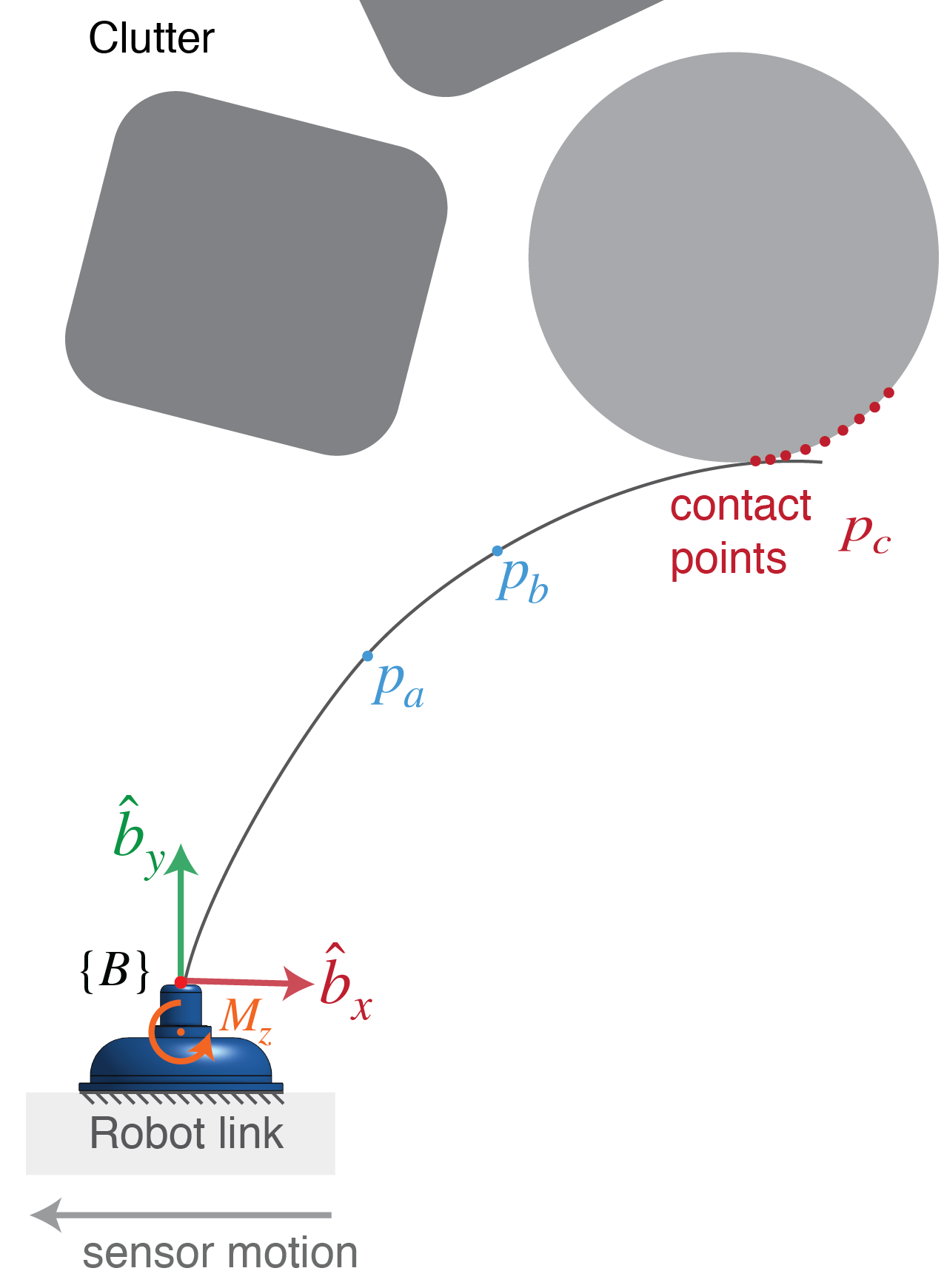}
  \caption{
  Figure illustrates a whisker sensor brushing along the surface of an object. The trace of the contact locations, $p_c$, over time informs the robot about the object surface.}
  \label{fig:working-principle}
\end{figure}

Details on the design, sensing principle and fabrication processes are provided in \cite{lin2022whisker}. Briefly, the whiskers consist of slender (0.2\,mm diameter 60 Y\,mm long) pre-curved nitinol wires. Each whisker is anchored to a compliant silicone rubber base that contains a small permanent magnet. A three-axis magnetometer on a PCB beneath the whisker measures the motions of the magnet to produce signals proportional to bending moments about the $\hat{b}_x$ and $\hat{b}_z$ axes, as depicted in \Cref{fig:working-principle}. 

\Cref{fig:working-principle} depicts an example of a whisker that has come into contact with a cylindrical object. As the robot moves in the 
$-\hat{b}_x$ direction, the whisker produces a moment $M_z$ at the sensor base that is measured continuously. This signal can be processed to generate two types of information: (i) proximity to nearby surfaces and (ii) locations of contacts. Proximity sensing enables safe navigation in constrained spaces by alerting the robot to nearby surfaces. Contact localization employs a Bayesian algorithm to map environmental contours during surface interaction. Together, these capabilities allow the robot to maneuver while also building a partial spatial reconstruction of nearby obstacles or surfaces.

In the following sections we present algorithms and models to enable these two types of feedback, and demonstrate their use cases. We show that proximity measurement can be processed in real-time from bending moments, and show how it can be used in a reactive controller to achieve obstacle avoidance when reaching in clutter. Simultaneously, sequences of bending moment measurements through time can be used in combination with known motion of the sensor base to probabilistically infer the locations of contacts. We show that a sequence of these contact location estimates can be used to reconstruct the shapes of surfaces that the sensor brushes against.

\section{Proximity sensing}
One of the challenges of operating a robot in unstructured environments with low visibility is being able to navigate spaces while avoiding collisions with objects. We avoid this difficulty by using sensors that have high structural compliance and also a large sensing volume or range. These properties allow the sensor to come into and remain in contact with objects while applying very low forces, and they provide ample buffer space to allow a robot to respond. 

We tested our whisker sensor and other commonly-used proximity sensors on different objects. We selected objects with various values of transparency and reflectivity, using a coffee mug as the baseline. In addition to our whisker sensor, we tested a VCNL4010 light sensor, a VL6180X laser sensor, and an HC-SR04 ultrasonic sensor. We calculated the Root-Mean-Square (RMS) error from the measured point to the nearest ground truth surface. Details on the measurement method and settings can be found in \Cref{sec:appendix}. As shown in \Cref{tab:compare-table}, the light and laser sensors are almost unusable on high-transparency objects (2,3). They also have a large noise on reflective 
objects (5-7). As the laser sensor emits a narrower infrared ray than the light sensor, it has a much larger error on the edges of highly reflective objects (7). While the ultrasonic sensor can measure the distance correctly with some noise, it has significant errors on non-flat surfaces (4,8) due to the wide range of its ultrasonic wave probe. Except for the rough rock (8), where local surface rugosity is about $1$\,mm, our whisker sensor achieves a distance measuring accuracy of less than 1\,mm.


\subsection{Safe navigation}
When non-zero deflection is measured by our sensors, it is a clear indication of the presence of an object within the sensing range of the whisker (\Cref{fig:geometry}). One implementation of a reactive controller can be to apply an artificial force at the sensor location in the direction counter to the normal, such the robot will move and avoid contact with the sensors attached. While this is a safe approach, we gain little information about the surroundings by fully avoiding contact with the whiskers. Instead, we are interested in allowing for sustained contacts with a safety boundary such that the controller will prevent objects from getting too close to the surface of the robot (\Cref{fig:system}). This approach affords more information gain by processing the contact location for shape reconstruction as will be shown in \Cref{sec:localization}, while also avoiding collisions with objects.

\begin{table}[t!]
\centering
\caption{RMS error
from sensor measurements to the nearest ground truth surface of different sensors and objects (units in mm). The VCNL4010, VL6180X, and HC-SR04 are light, laser, and ultrasonic time-of-flight sensors, respectively. Objects 2,3 have high transparency; objects 4-8 have varying levels of reflectivity. Object 1 (non-transparent, matte finish) is used as a baseline.}
\label{tab:compare-table}
\resizebox{0.49\textwidth}{!}{%
\begin{tabular}{>{\centering\arraybackslash}m{2cm}cccccccc}
\toprule
& 
\includegraphics[width=0.95cm,height=0.95cm]{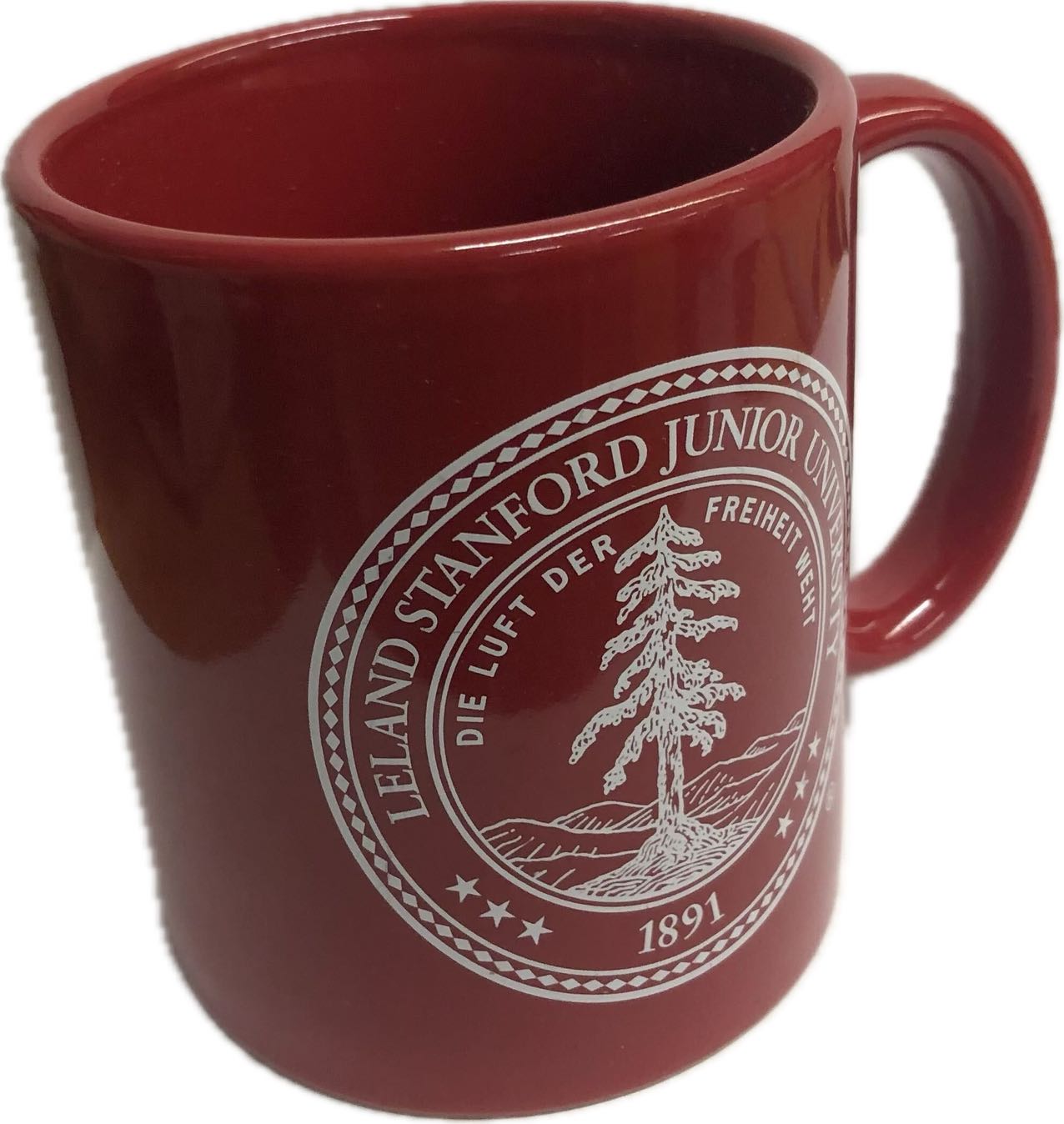} & \includegraphics[width=0.5cm,height=1cm]{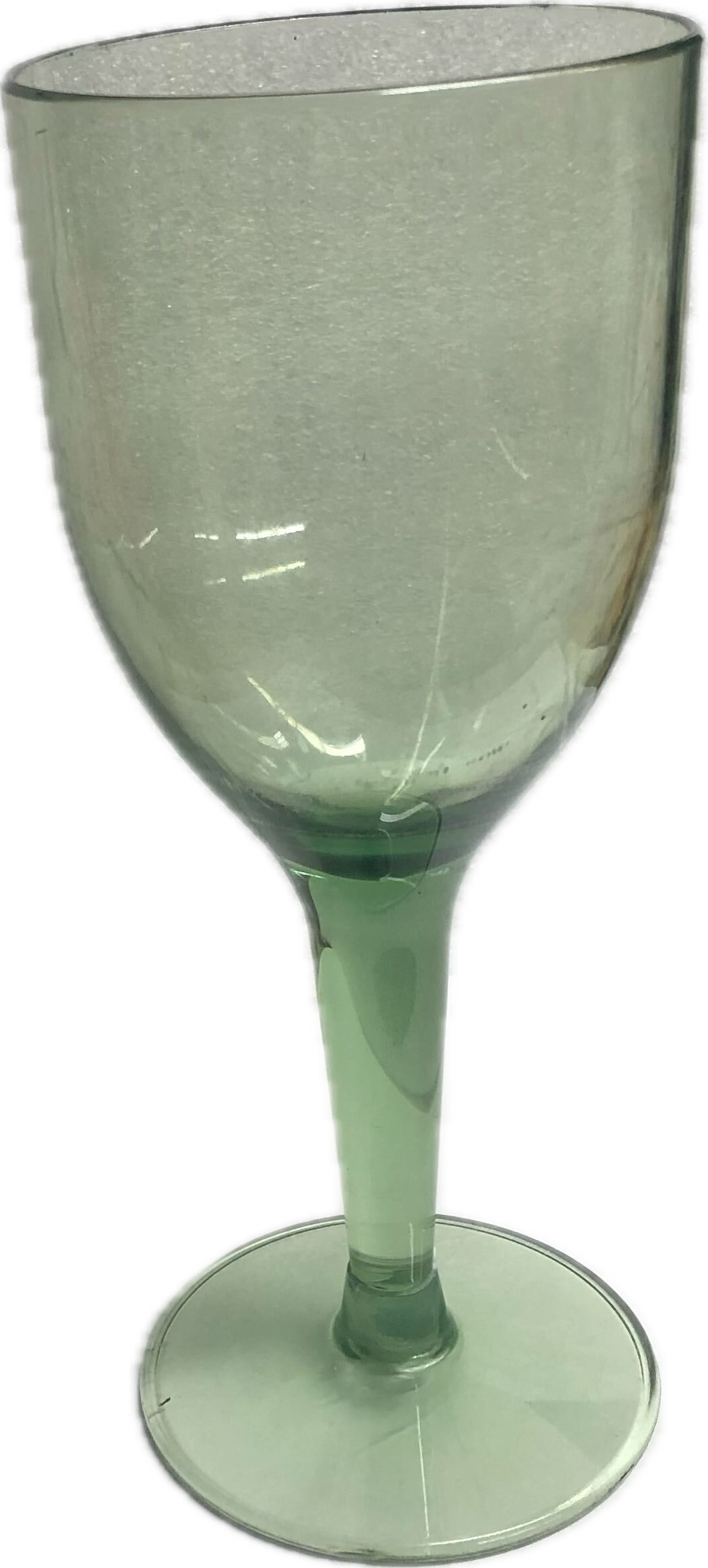} & 
\includegraphics[width=0.5cm,height=1cm]{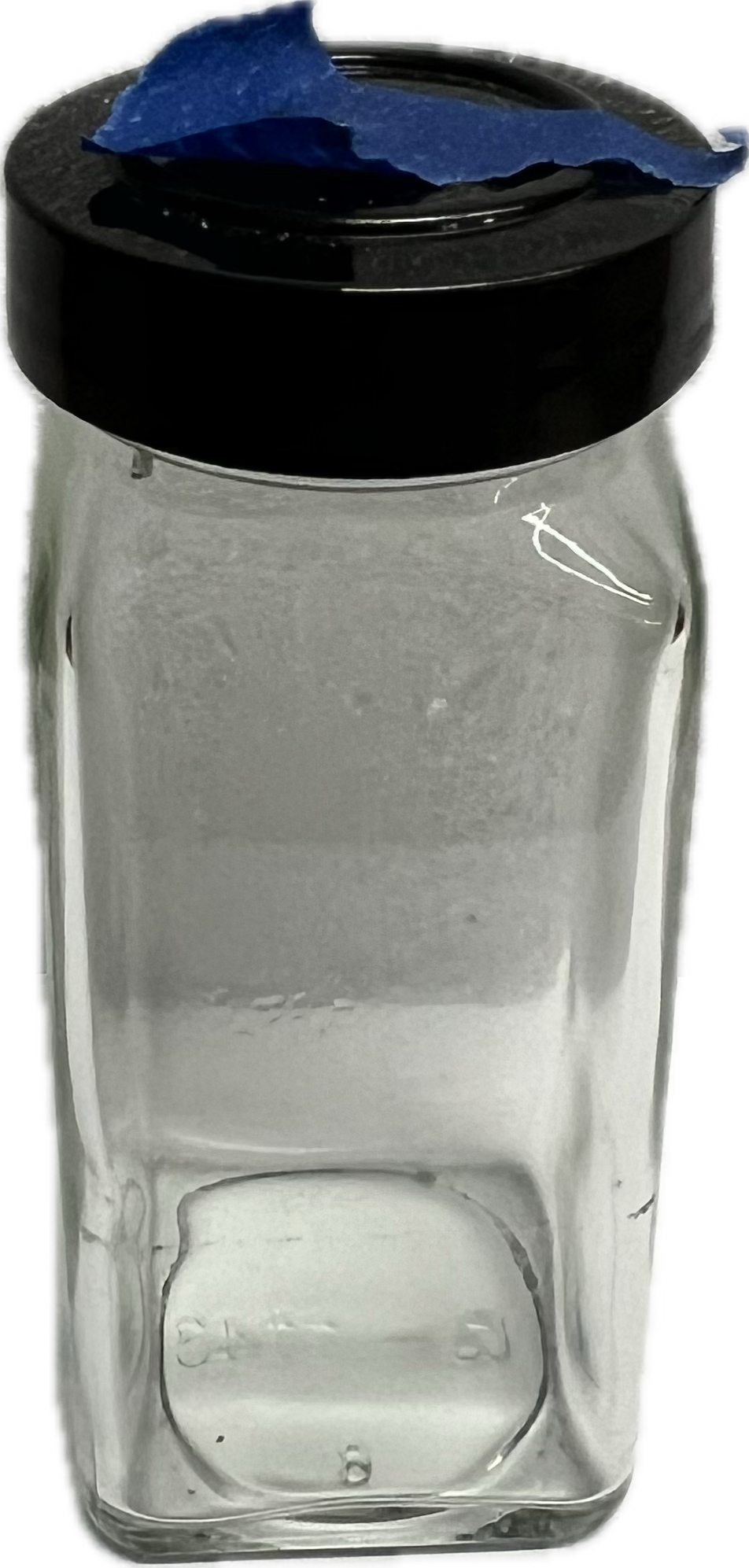} & \includegraphics[width=0.5cm,height=1cm]{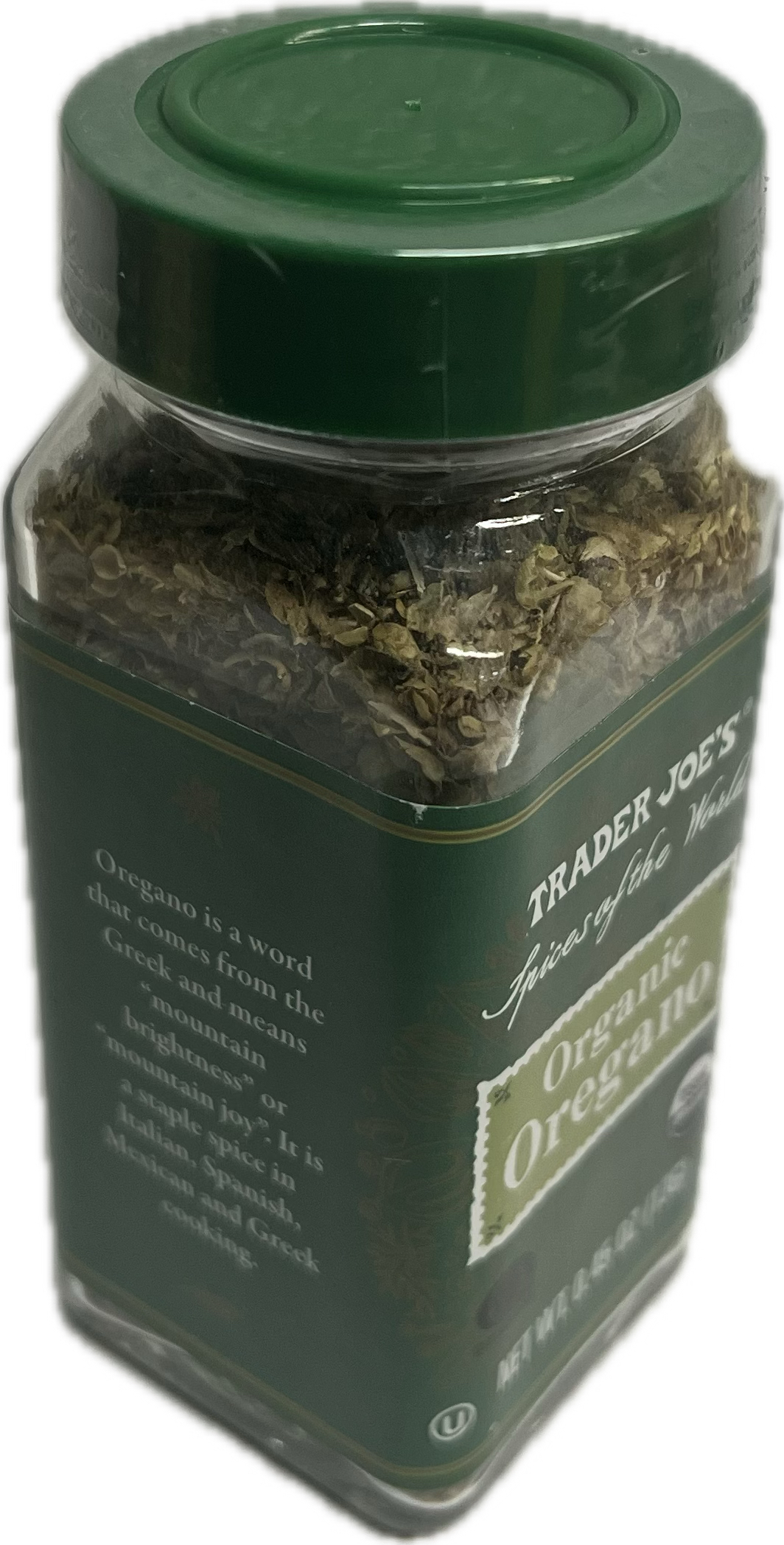} & \includegraphics[width=0.5cm,height=1cm]{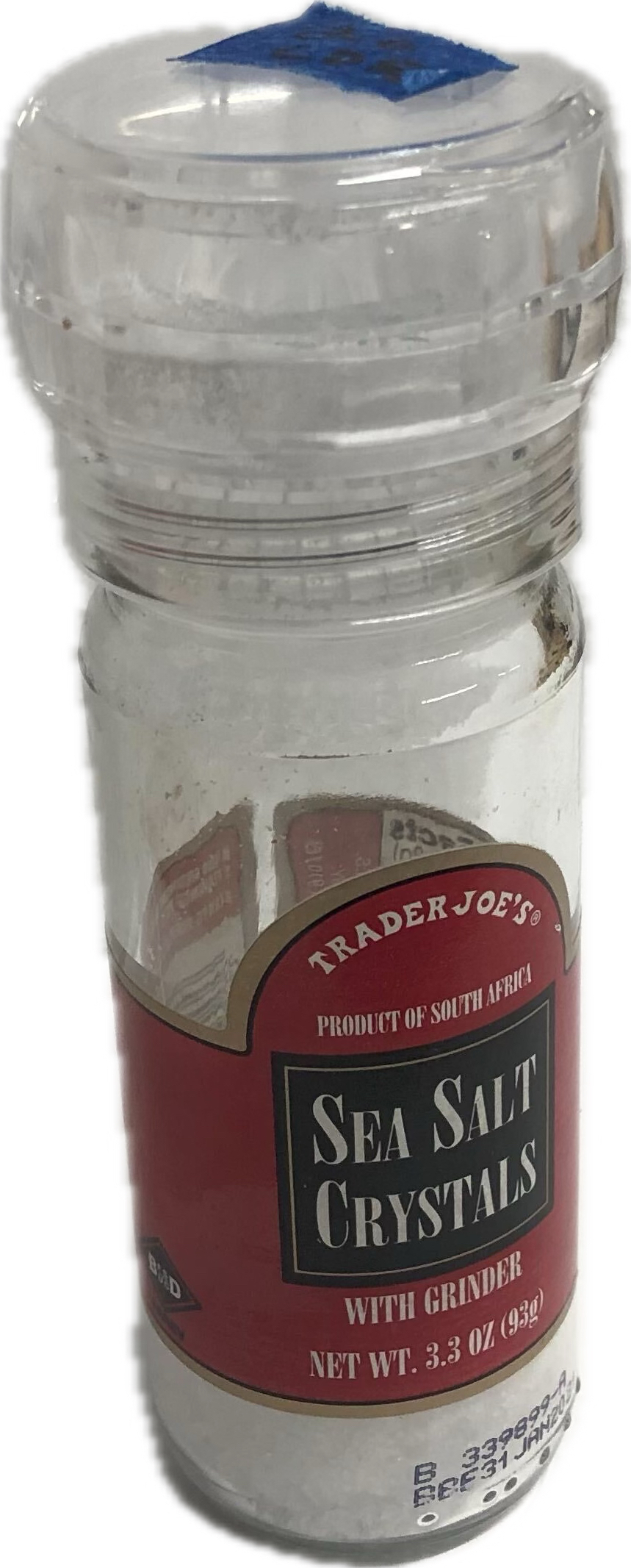} & 
\includegraphics[width=0.7cm,height=1cm]{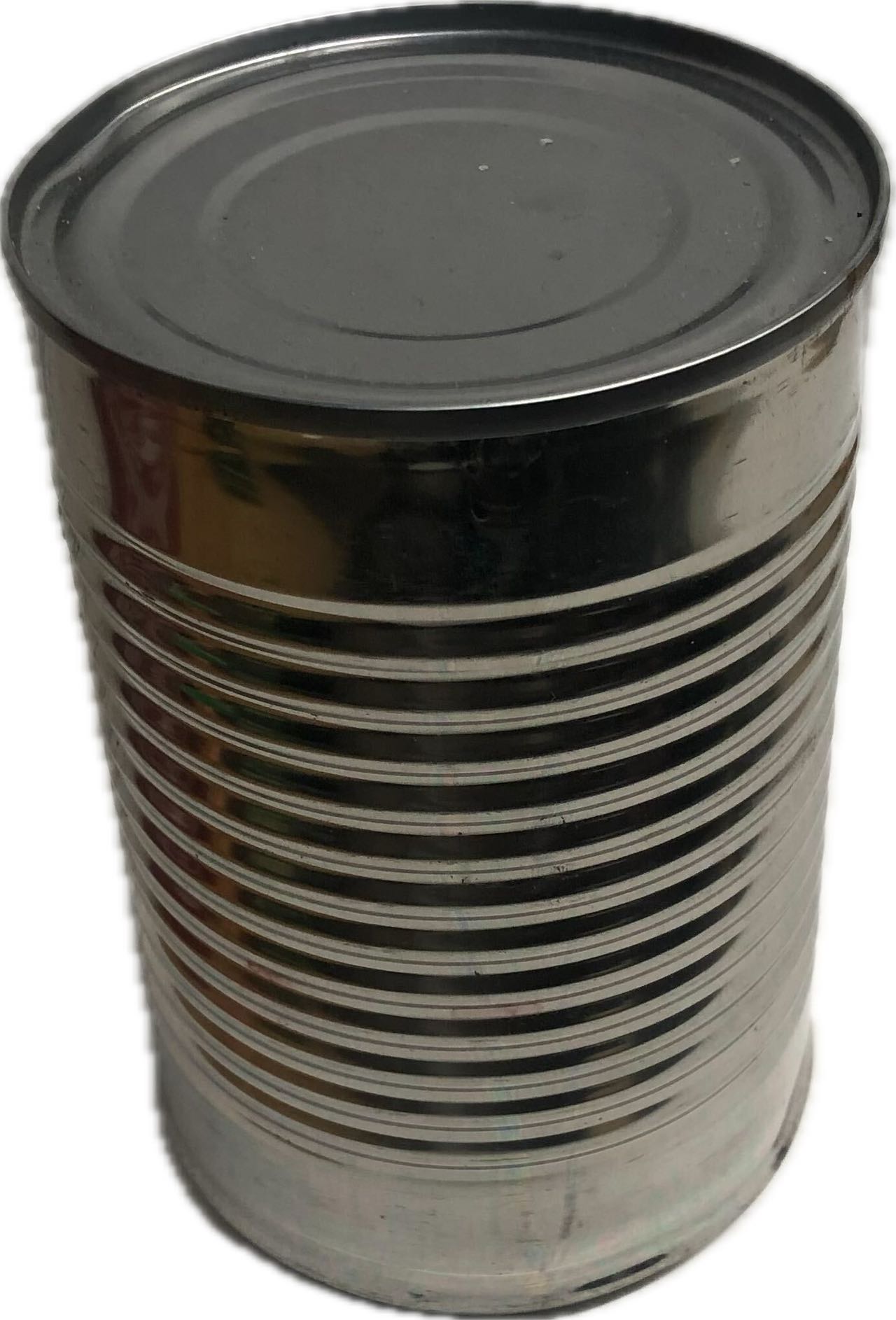} & \includegraphics[width=1cm,height=0.8cm]{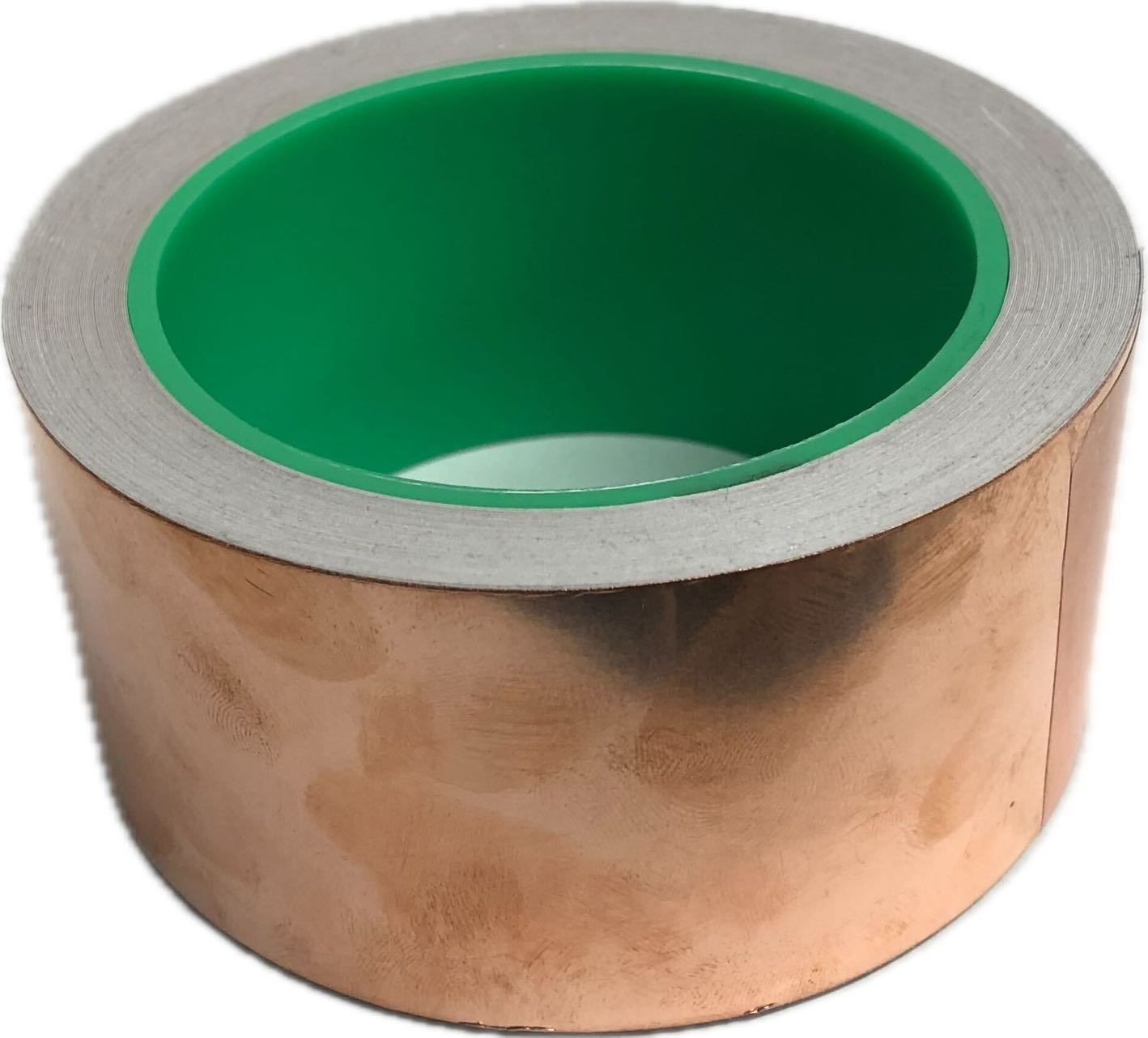} & \includegraphics[width=1 cm,height=0.8cm]{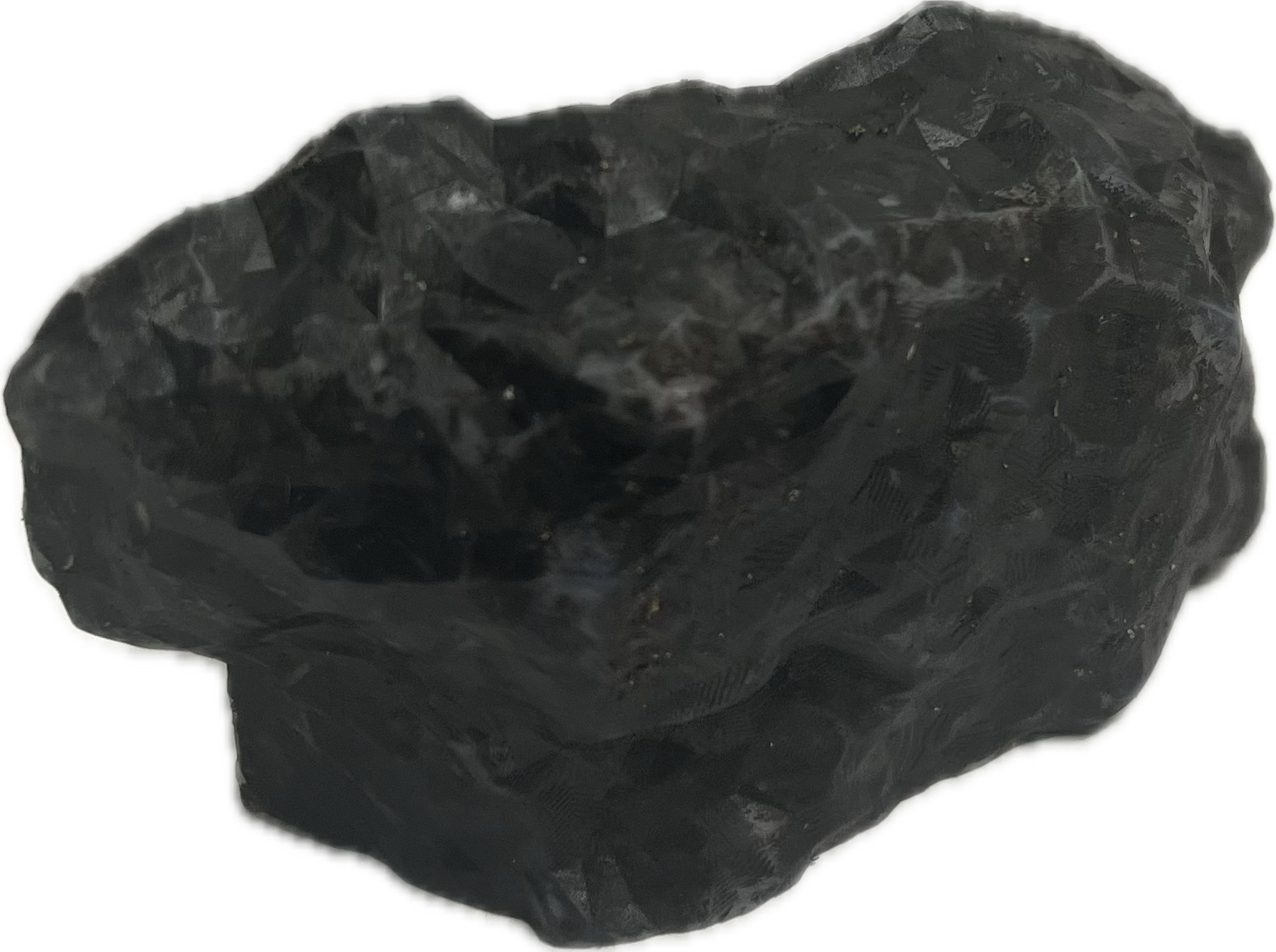} \\
\midrule
 VNCL4010       & 1.99   & 10.29  & 8.83   & 1.59      & 8.17   & 2.38   & 3.34   & 15.36  \\
    VL6180X & 2.38  & 18.19 & 11.14 & 5.71 & 8.09  & 7.45  & 11.90 & 8.83  \\
    HR-SR04 & 2.92  & 2.13  & 2.25  & 11.18  & 3.17  & 3.17  & 2.31  & 15.99 \\
    Whisker & 0.16 & 0.60 & 0.75 & 0.95 & 0.44 & 0.32 & 0.29 & 1.29 \\
\bottomrule
\end{tabular}%
}
\end{table}

We implemented a reactive collision avoidance controller using a admittance-type control (similar to that found in \cite{lynch2017modern}): 
\begin{align}
F^*&=-k_v\dot{x}+k_p(x_g-x)\\
\dot{x}_g&=\sum_{i=0}^{N} \prescript{B}{E}{J^T_i}F_{r,i}\\
F_{r,i}&=\gamma\max(M_{z,i}-M_{z,th}, 0)
\end{align}
where $x$ is the end-effector Cartesian pose, $x_g$ is the goal pose, $\prescript{B}{E}{J^T_i}$ is the Jacobian that relates forces at the i-th sensor reference frame to forces and moments at the end-effector frame, $M_z,i$ is the deflection measured at the i-th sensor (in this case measured in sensor counts), $M_{z,th}$ is a user defined threshold for measured deflection at which artificial forces start acting, finally $\gamma$ is a scalar found empirically that relates these deflections to the rate at which the controller moves away from the penetrating surface.

We implemented this reactive controller on the system shown in \Cref{fig:system} comprised of a Flexiv Rizon robot arm equipped with an end-effector covered with an array of 16 whisker sensors. We controlled the robot to move among a clutter of objects at moderate speeds (4\,cm/s). We tested this controller under three different scenarios:
\begin{enumerate}
    \item Scenario 1: Robot moves at full speed towards a wine glass.
    \item Scenario 2: Robot is teleoperated in arbitrary motions into collision with a clutter of objects.
    \item Scenario 3: Robot is commanded to move with a constant velocity vector towards a clutter of objects with an opening of free-space.
\end{enumerate}
Success was measured as not toppling objects over or shifting them from their resting position. We performed 5 experiment trials using Scenario 1, where the robot was commended at a speed of 4\,cm/s towards a wine glass. In all 5 trials, the controller was able to push the robot away from the collision upon sensing contact with the whiskers. For Scenario 2, we used a joystick to teleoperate the motion of the robot intentionally moving towards collision with a clutter of objects. Throughout the experiment, which lasted for 40 seconds, the controller was able to avoid all hard collisions. We performed 4 trials using Scenario 3 with different object arrangements. Light object collisions were observed in two trials but caused no toppling or appreciable object shifting, and in all four experiments the controller was able to guide the robot through the free-space opening among objects. The result of this collision avoidance behavior is best seen in the accompanying video ~\footnote{Reactive collision avoidance experiments video\url{https://youtu.be/Rvx4tWSkfu8}}.

\section{Contact localization}
\label{sec:localization}

Contacts on the whisker sensor cause bending rotations that reflect as bending moments at the base. However, there exists infinite contact locations that can result in the same whisker deflection and corresponding moments at the base. With just instantaneous moment measurements, contacts at points $p_c$, $p_a$ and $p_b$ in \Cref{fig:working-principle}, would be indistinguishable.
Thus, to obtain information on the contact location along the whisker, we integrate instantaneous bending rotations at the whisker base due to whisker deflection, as well as motion of the whisker base over time. The insight is that given multiple moment measurements at different sensor locations---and assuming that the contact location remains mostly the same in the world reference frame---it is possible to infer the contact location from the sequence of moment and position measurements. In this section we detail our approach using these measurements and Bayesian filtering algorithms to infer contact point location and motion.

\subsection{Contact localization model and assumptions} \label{sec:contactModel}

We begin by summarizing the assumptions made to limit the scope of the problem.
\begin{enumerate}
\item Objects that come into contact are immobile in the world reference frame. In other words, if the robot does not move then contact locations do not change.
\item There is at most one contact point on a whisker with the environment at any time. This will be true for convex objects.
\item Frictional forces along the whisker are negligible in terms of their effect on the sensor. This again will generally be true given the nitinol whisker material.
\end{enumerate}

In the following derivation we follow the conventions in~\cite{murray2017mathematical} with lowercase boldface fonts for vectors, uppercase boldface fonts for matrices and $\{F\}$ denoting a reference frame. Lowercase superscript and subscript letters indicate the reference frames in which a vector or matrix is expressed. For example, $\bm{v}^a_{ab}$ is the linear velocity of reference frame A relative to B (subscript ${ab}$) as viewed in reference frame A (superscript $a$).

Let $\{B\}$ be the reference frame of the sensor base and $\{S\}$ be the spatial (world-fixed) reference frame. We define our process and sensor model as follows
\begin{align}
    \bm{x}_{k+1} &= \bm{A}\bm{x}_k+\bm{B}\begin{bmatrix}\bm{v}^b_{sb}\\\bm{\omega}^b_{sb}\end{bmatrix}+\bm{w}_k\label{eq:procModel}\\
    y_k &= g(\bm{x}_k)+\bm{\nu}_k
\end{align}
Where $\bm{x}_k=\bm{p}_c$ is the position vector of the contact point relative to the origin of $\{B\}$ at time-step $k$. $y_k$ is the sensor measurement at time-step $k$. $\bm{v}^b_{sb}$ and $\bm{\omega}^b_{sb}$ are the linear and angular velocity, respectively, of sensor base $\{B\}$ relative to world-fixed frame $\{S\}$ as viewed in the body frame. Process and sensor noise are modelled as Gaussian white noise where $\bm{w}_k, \bm{\nu}_k\sim \mathcal{N}(0,\,\sigma^{2})$. $g:\mathbb{R}^2 \to \mathbb{R}$ is our sensor model that maps contact location to moment measurements which we will elaborate on in \Cref{sec:calib}.


To define the process model, we first find the velocity of point $p_c$ relative to $\{S\}$ as viewed in the body frame $\{B\}$ denoted as $\bm{v}^b_{sp_c}$. Let's consider $p_c$ as a point with respect to $\{B\}$, and define a point $b_p$ that is instantaneously coincident to $p_c$, but fixed in $\{B\}$. We can find the linear velocity of $p_c$ with respect to the spatial frame as
\begin{align}
    \bm{v}^b_{sp_c} &= \bm{v}^b_{sb_p} + \bm{v}^b_{bp_c}\\
                    &= \bm{v}^b_{sb} + \bm{\omega}^b_{sb}\times \bm{b}_p + \bm{v}^b_{bp_c}
\end{align}
where $\bm{v}^b_{sb_p}$ is the linear velocity of point $b_p$ relative to $\{S\}$ as viewed in the body frame, $\bm{v}^b_{sb}$ is the linear velocity of $\{B\}$ relative to $\{S\}$ as viewed in the body frame, $\bm{\omega}^b_{sb}$ is the angular velocity of $\{B\}$ with respect to $\{S\}$. $\bm{p}_c$ is the position vector of point $p_c$ relative to $\{B\}$'s origin.

Given our assumption that the contact point remains static in the spatial frame (i.e. $\bm{v}^b_{sp_c}=\bm{0}$), the contact point velocity relative to $\{B\}$ is
\begin{align}
    \bm{v}^b_{bp_c} &= -\bm{v}^b_{sb} + -\bm{\omega}^b_{sb}\times \bm{b}_p\\
                    &= -\begin{bmatrix}\bm{I} & [\bm{p}_c]\end{bmatrix}\begin{bmatrix}\bm{v}^b_{sb}\\\bm{\omega}^b_{sb}\end{bmatrix}
\end{align}
where $[\bm{p}_c]$ is the skew-symmetric matrix of vector $\bm{p}_c$ and $\bm{I}$ is the identity matrix. 

As the sensor is attached on a robot link, $\bm{v}^b_{sb}$ and $\bm{\omega}^b_{sb}$ can be found through 
\begin{align}
    \begin{bmatrix}
\bm{v}^b_{sb}\\\bm{w}^b_{sb}
\end{bmatrix}=\bm{J}^b_{sb}(\bm{q})\bm{\dot{q}}
\end{align}
where $\bm{J}_{sb}^b$ is the body Jacobian of $\{B\}$ relative to $\{S\}$, and q is the vector of manipulator joint angles.

With the velocity of the contact point with respect to $\{B\}$ defined, we can express the process model as
\begin{align}
    \bm{x}_{k+1} &= \bm{x}_k + \delta_t\bm{v}^b_{sp_c} = \bm{x}_k -\delta_t\begin{bmatrix}\bm{I} & [\bm{p}_c]\end{bmatrix}\bm{\xi}^b_{sb}+\bm{w}_k \label{eq:predict-step}
\end{align}
which gives as $\bm{A}=I$ and $\bm{B}=-\delta_t\begin{bmatrix}\bm{I} & [\bm{p}_c]\end{bmatrix}$ in \Cref{eq:procModel}.

In the case when contacts are not static, such as when they travel along objects having a large radius of curvature, the velocity of the contact point relative to the world $\bm{v}^b_{sp_c}$ is non-zero. However, since we assume no prior knowledge of the object shape and location, the contact velocity is not predictable. As inferences are executed very fast (250\,Hz) contact points do move far between iterations, so we choose to make the static assumption and correct for errors using the sensor model.

\subsection{Sensor model and calibration} \label{sec:calib}
\begin{figure}[htb!]
  \centering
  \includegraphics[width=0.85\linewidth]{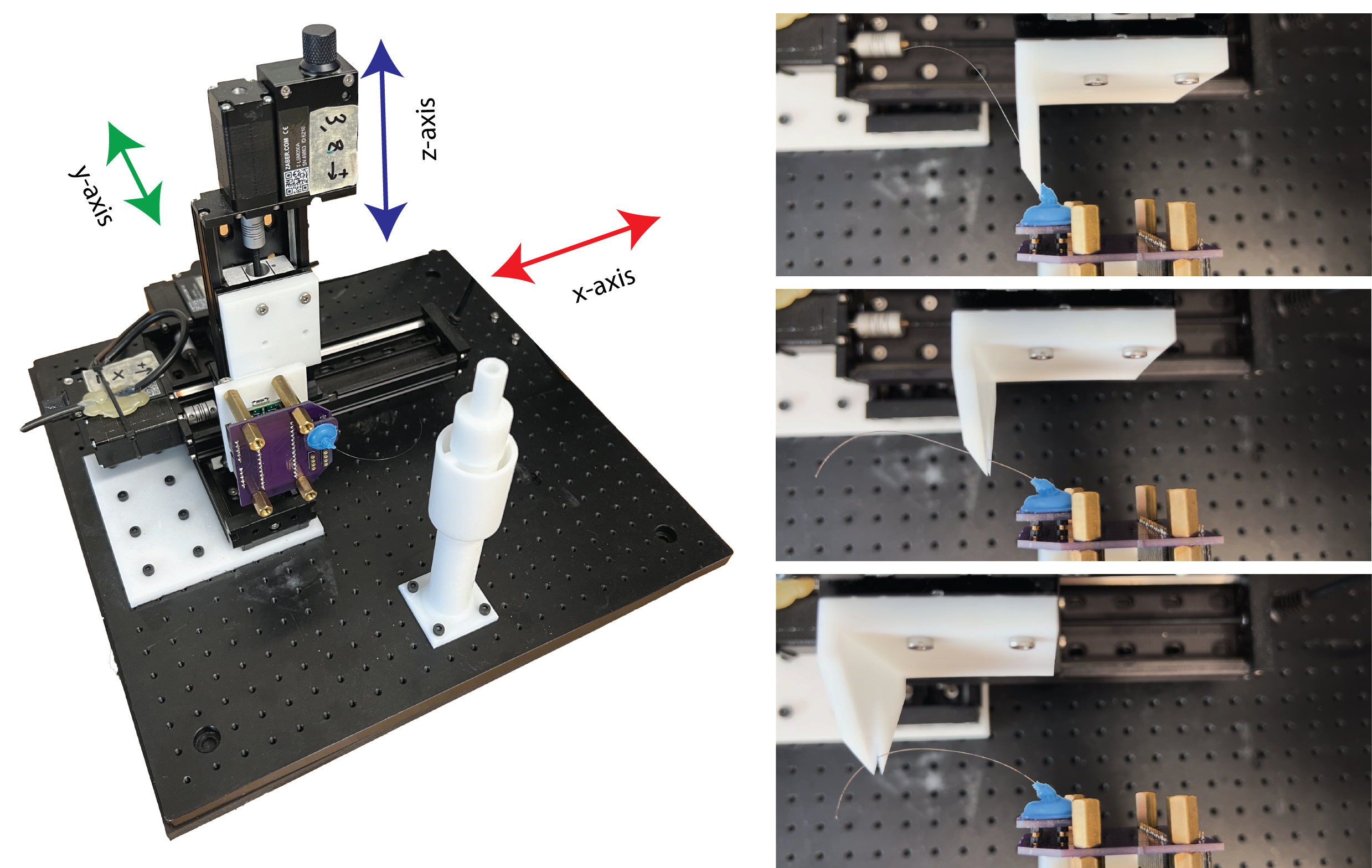}
  \caption{Left: 3-axis positioning stage used for calibrating whisker sensors. Right: Example of different calibration points collected within the sensing region.}
  \label{fig:calibration-stage}
\end{figure}

In order to find the sensor model, $g_x$ and $g_y$, which maps contact point locations, $\bm{x}_k$, to predicted moment measurements, $\bm{y}_k$, we develop a calibration platform and procedure. To gather data for this mapping we used a 3-axis calibration stage shown in \Cref{fig:calibration-stage}: model LSM100A from Zaber (position resolution of $0.1 \mu$m). A calibration rig is attached at the end of the stage which has a V-shaped groove used to enforce a known contact location on the whisker. During data collection, position and magnetic sensor data are recorded as the end-effector of the stage is driven to deflect the whisker with the calibration rig. This procedure is then repeated for different positions of the stage by tracing an arbitrary trajectory (as shown on the right of \Cref{fig:calibration-stage}) that spans the sensing region.

From calibration we obtain a dataset $\mathcal{D}=\{(\bm{x}_i,\bm{y}_i)\mid i=1,...,n\}$ where $\bm{x}_i$ are 3D positions of the contact location and $\bm{y}_i$ is the bending moment measurements from the Hall effect sensor in x and y axes as illustrated previously in \Cref{fig:working-principle}. 

We use the method of Gaussian Process Regression (GPR) to find a model that will best fit the calibration dataset that maps three-dimensional position input to two-dimensional sensor response output. We use GPR as it can yield highly expressive models and works well for noisy observations. 

The underlying function that relates the data points is a kernel function. Many alternative kernel functions also exist \cite{williams2006gaussian}, and in practice, kernels are chosen through empirical results or through prior knowledge about the underlying distribution. The most frequently used kernel function is the Squared Exponential kernel, also known as the Radial Basis Function (RBF). In this work, we found that a Thin-plate (TP) kernel yielded a sensor model that is more robust to tracking divergence compared to models obtained through the RBF kernel. The two kernels are defined as follows:
\begin{align}
    k_{SE}(r) &= \exp{(\frac{-r^2}{2l^2})}\\
    k_{TP}(r) &= \frac{1}{12}(2|r|^3 - 3Rr^2 + R^3)
\end{align}
where $l$ is a hyperparameter that tunes the locality of the RBF (the higher $l$ is, the less influence data points have on farther neighbors), and $R$ is the radius of the region within which the TP model will minimize the second-order gradient.

\begin{figure}[htb!]
  \centering
  \includegraphics[width=\linewidth]{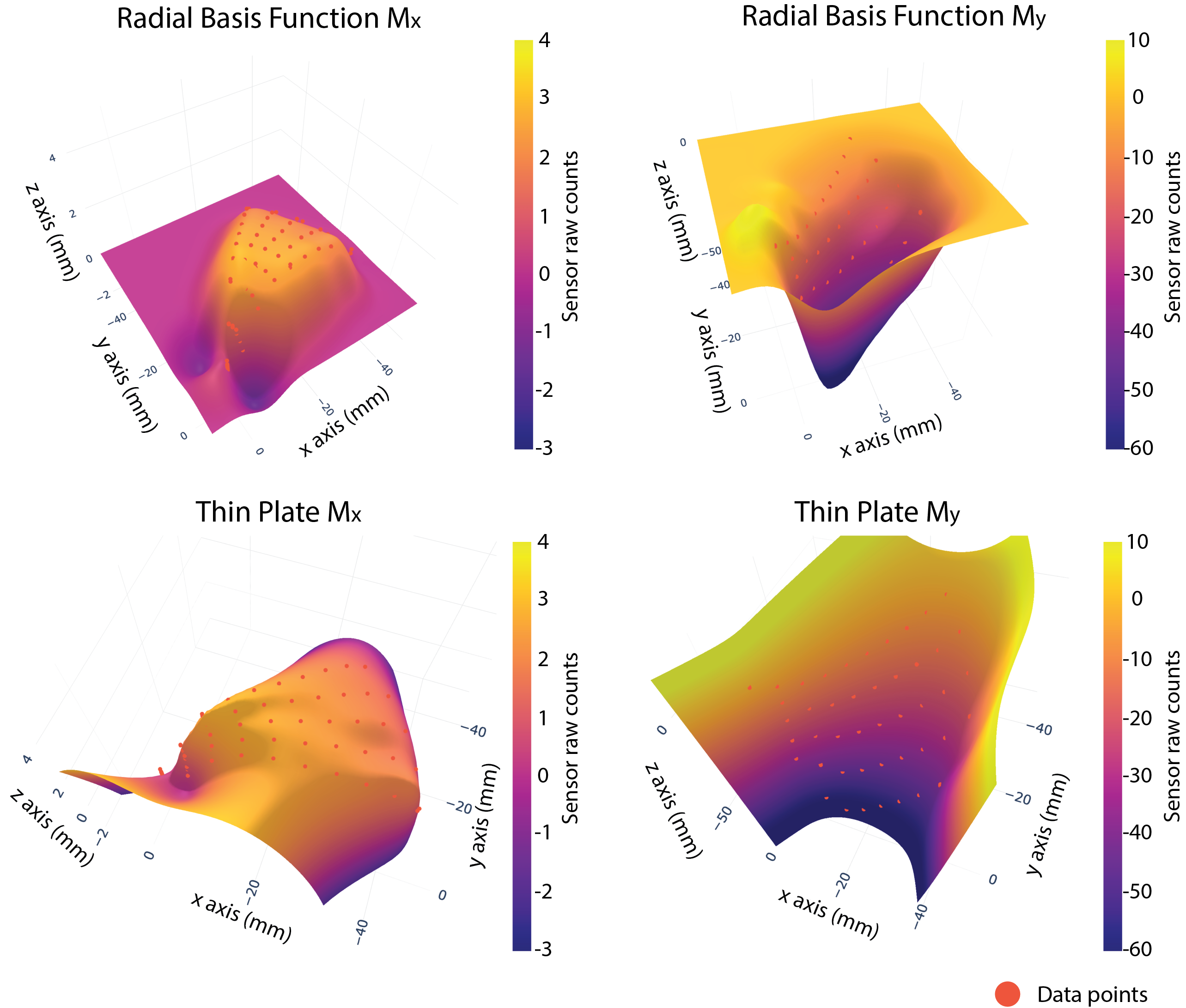}
  \caption{Sensor data fitted with Gaussian Process Regression. Top plots use Radial Basis Function kernel. Bottom plots use Thin-plate kernel.}
  \label{fig:sensor-model}
\end{figure}

One notable difference between models from a TP kernel and RBF kernel is that at the extrapolated region (i.e., the domain outside of where data was collected) the TP kernel maintains a mean with a constant slope whereas RBF has a mean that converges to zero \cite{williams2006gpis}. This can be visualized in \Cref{fig:sensor-model}, where we show a surface plot of a slice of the model at $z=0$. From the RBF models we can see that the mean value converges to zero exponentially at a distance away from data points, whereas, TP models have a constant slope that extends from the edges of the data cluster. This is because the TP kernel is formulated to minimize the second-order gradient \cite{williams2006gpis}.

Models from both kernels fit the dataset very well. The R-squared values for RBF (with hyper-parameter $l=5$) are $0.99815$ and $0.99770$ for $M_x$ and $M_y$, respectively, and values for TP (with hyper-parameter $R=100$) are $0.99982$ and $0.99969$ for $M_x$ and $M_y$, respectively. However, it is more advantageous to use the TP models because their extrapolation behavior yields a more stable tracking. This will be further elaborated in ~\Cref{sec:sensor-model-comparison}.

\subsection{Sensor model comparison}
\label{sec:sensor-model-comparison}

\begin{figure}[htb!]
  \centering
  \includegraphics[width=\linewidth]{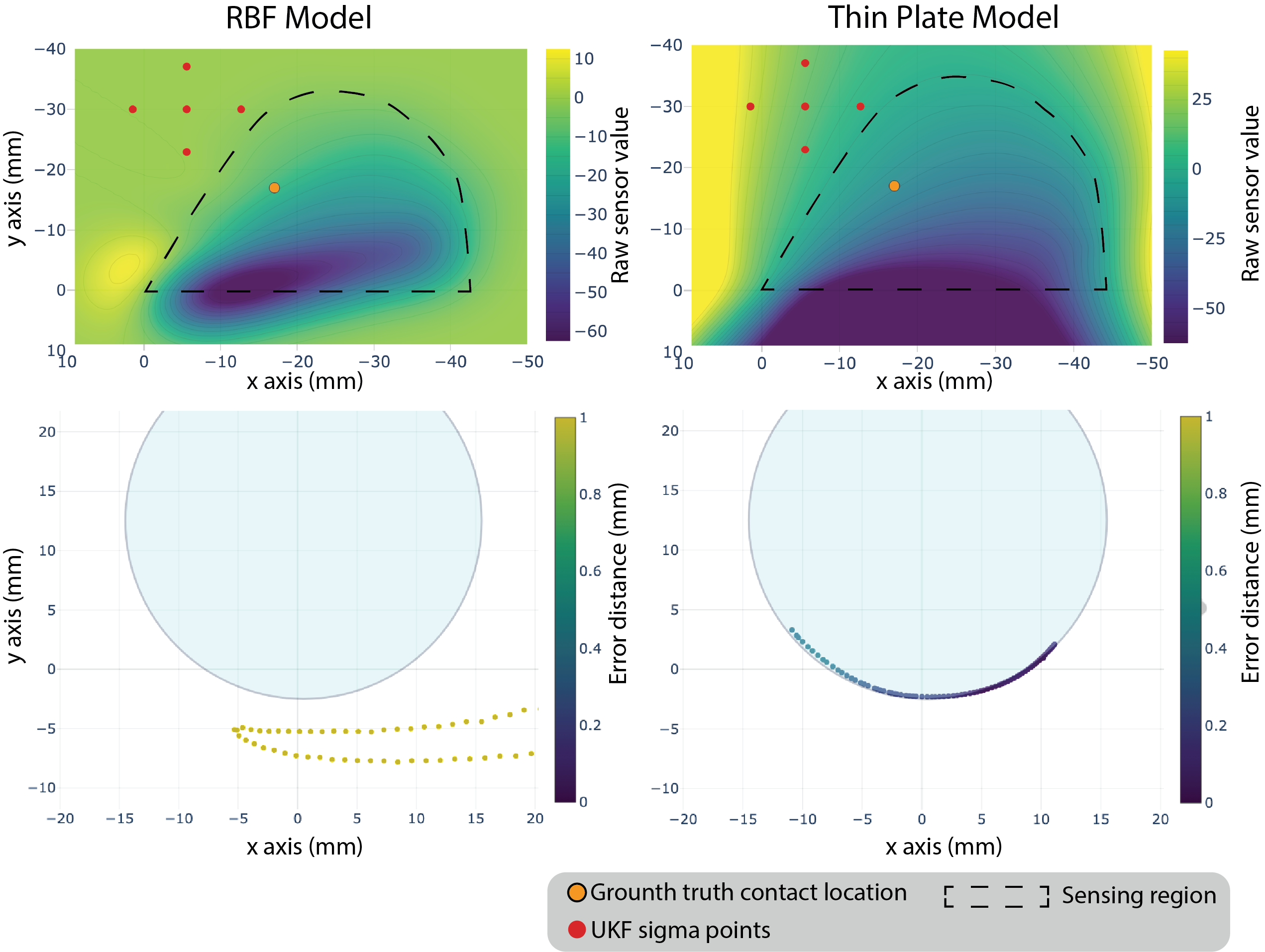}
  \caption[Contact localization methods]{Contact localization example for demonstrating the advantage of the Thin Plate (TP) model over RBF model. Left top shows the initial condition where the UKF sigma points are initialized outside of the sensing region. The ground truth location is within the dashed region. Left bottom shows the tracking results using the RBF model where divergence occurs due to extrapolation that is unstructured. Right top shows the initial conditions overlaid on the TP model. Right bottom shows the tracking results converge and trace the contour of the object correctly.}
  \label{fig:sensor-model-comaprison}
\end{figure}

As mentioned in \Cref{sec:calib}, models obtained from the TP kernel are favorable for contact localization. An example of this is illustrated in \Cref{fig:sensor-model-comaprison} where we perform a contact localization experiment with a semi-curved sensor (as illustrated in \Cref{fig:geometry}) brushing against a cylinder object. We run the tracking algorithm presented in the previous section with the same data and initial conditions, but in one case using the RBF sensor model and the other case using the TP sensor model. In the top two plots we show, overlaid on the sensor models, the goal contact location in green and the UKF sigma points in red. The initialization or prior is such that the sigma points lie in the extrapolated region. It can be observed that, in the RBF model case, the gradients of the region covered by the sigma points are much flatter and the gradient direction might not provide a good correction during the Measurement Update step \cite{wan2000unscented}. In comparison, the gradients in the TP model case are more directed and pronounced in the extrapolated region, making it faster in the Update step to converge towards the correct state. The two bottom plots in \Cref{fig:sensor-model-comaprison} then show the different results where the RBF model diverges and TP model converges quickly even though both have the same initial condition and measurement data.

Apart from the empirical data showing that the TP model provides more stable tracking, another reason to use the TP model is that its extrapolated data make more sense than for RBF. The reason the calibration data points are bounded to a region is that the whisker sensor has a finite length, and large deflections cause the calibration rig to break contact with the sensor wire. However, if the whisker had infinite length then we would expect the sensing range to be extended expecting the sensor signal to continue the trend at the boundaries. Given this behavior, the TP model with a constant slope away from the boundaries is a more accurate approximation compared to the RBF model that converges to a zero mean.

\subsection{Contact location inference with Bayesian Filtering} \label{sec:bayesian}
In summary, the steps for localization are:
\begin{enumerate}
    \item Make an initial guess $\mu_0$ and $\Sigma_0$
    \item Process proprioception data to perform a prediction step as in \Cref{eq:procModel}.
    \item Correct for state distribution based on the sensor prediction.
\end{enumerate}

To track contact locations from a sequence of base moment measurements, we use Bayesian filtering in a recursive algorithm that infers the state distribution from a history of sensor data and control inputs. The posterior distribution is defined as
\begin{align}
    b(\bm{x}_t) &=p(\bm{x}_t \mid \bm{y}_{1:t}, \bm{u}_{1:t}) \\
    &= \eta \; p(\bm{y}_t\mid \bm{x}_t,\bm{u}_t)\int p(\bm{x}_t\mid \bm{x}_{t-1},\bm{u}_t)b(\bm{x}_{t-1})d\bm{x}_{t-1}
\end{align}
where $\eta$ is a normalization factor, $b(\bm{x}_{t-1})$ is the prior distribution, and $p(\bm{y}_t\mid \bm{x}_t,\bm{u}_t)$ and $p(\bm{x}_t\mid \bm{x}_{t-1},\bm{u}_t)$ can be obtained from the sensor and process model, respectively.

We implemented three different non-linear Bayesian filters to compare their performance: Extended Kalman Filter (EKF), Unscented Kalman Filter (UKF) and Particle Filter (PF). A process noise covariance of $1e^{-5}I_2$ was empirically found to work well for EKF and UKF. This low noise is reasonable given the accuracy of the optical encoders and calibration system. The sensor noise variance was chosen to be 0.25 which was found through the RMSE of the calibration results reported previously. The process and sensor noise covariance for PF $1e^{-3}I_2$ and 1 respectively, which were also found empirically to track with low error. We used N=1000 particles for PF.

As mentioned previously, when the actual contact location travels along the surface of an object, our process model described in \Cref{sec:contactModel} will be inaccurate, which may lead to divergence of the estimate. We cannot model this behavior but we can use known techniques for compensating for model errors. These include adding fictitious process noise or using a Fading Memory (FM) filter \cite[p.\,139]{simon2006optimal}. Both are used to increase the predicted covariance but do so differently. FM scales the prior covariance while fictitious process noise adds a constant positive variance to the diagonals. FM was empirically found to produce better results in our application. We implemented this method on the Bayesian filters by scaling the prior covariance by a factor $\alpha=1.004$ at every time step. 


\begin{algorithm}[htp!]
    \caption{Unscented Kalman Filter for Contact Tracking}\label{alg:ekf}
  \begin{algorithmic}
  \Procedure{Unscented\_Kalman\_Filter}{$\mu_{t},\Sigma_{t}, u_t, z_t$}
    \If{mod(k, $T_{\mathrm{infl}}$) == 0}
        \State $\Sigma_{t} \gets \gamma_{\mathrm{infl}}\Sigma_{t}$
    \EndIf
    \State $\chi_{t-1} = \left( \mu_{t-1} \quad \mu_{t-1} + \gamma \sqrt{\Sigma_{t-1}} \quad \mu_{t-1} - \gamma \sqrt{\Sigma_{t-1}} \right)$
    \State $\chi_{t}^* = g(u_t, \chi_{t-1}) \quad \mu_t = \sum_{i=0}^{2n} w_m^{[i]} \chi_{t}^{*[i]}$ 
    \State $\Sigma_t = \sum_{i=0}^{2n} w_c^{[i]} \left( \chi_{t}^{*[i]} - \mu_t \right) \left( \chi_{t}^{*[i]} - \mu_t \right)^T + R_t $
    \State $\hat{\chi}_t = \left( \mu_t \quad \mu_t + \gamma \sqrt{\Sigma_t} \quad \mu_t - \gamma \sqrt{\Sigma_t} \right)$
    \State $\hat{z}_t = h(\hat{\chi}_t) \quad \bar{z}_t = \sum_{i=0}^{2n} w_m^{[i]} \hat{z}_{t}^{[i]}$
    \State $S_t = \sum_{i=0}^{2n} w_c^{[i]} \left( \hat{z}_{t}^{[i]} - \bar{z}_t \right) \left( \hat{z}_{t}^{[i]} - \bar{z}_t \right)^T + Q_t$
    \State $\Sigma_{t}^{z} = \sum_{i=0}^{2n} w_c^{[i]} \left( \chi_{t}^{*[i]} - \mu_t \right) \left( \hat{z}_{t}^{[i]} - \bar{z}_t \right)^T$
    \State $K_t = \Sigma_{t}^{z} S_t^{-1} \quad \mu_t = \mu_t + K_t(z_t - \bar{z}_t)$
    \State $\Sigma_t = \Sigma_t - K_t S_t K_t^T$
  \EndProcedure
  \end{algorithmic}
\end{algorithm}

In addition to implementing these filters, we compared tracking performance to a baseline by Solomon \emph{et al.}~\cite{solomon2010extracting}. Similar to our method, this baseline estimates contact points at every time step but with estimates that are deterministic rather than probabilistic (i.e., single values rather than a distribution). Another distinction between our method and this baseline is that our sensor model is calibrated, while the baseline uses a small deflection Euler-Bernoulli beam model. In our implementation of this baseline, we used our calibrated sensor model to obtain predicted moments ($M_{i+1}$) and the arc-length to torque ratio ($\frac{ds}{M_i}$) which are used for the estimation correction step. We compare our localization methods to the baseline only in the straight whisker case, as the baseline assumes a nominally straight beam and it would take significant effort to adapt it to an arbitrary shaped beam. For more detail on the baseline please refer to \cite{solomon2010extracting}.

\subsection{Reusing history of contact points}
\label{sec:reusing-contacts}
As detailed in \Cref{sec:bayesian}, the localization method is a recursive algorithm, and how quickly it converges to the ground truth depends on how good the prior distribution is. Since each interaction between whisker and objects can be brief, it is important that the algorithm can converge quickly.

We do not assume we have any prior information about the world, so it is only possible to guess where contacts will happen on the whisker sensor. However, as information is collected from the environment, we can integrate this data into a map that can then be used to make a more accurate guess on the prior. To build this map we use Bayesian Hilbert Maps (BHM) developed by Senanayake et al. \cite{senanayake2017bayesian} which is a method for building continuous occupancy maps in static environments. Two types of data are collected: (i) occupied points which are the mean estimates from the contact localization method and (ii) unoccupied points which are randomly sampled from within the volume occupied by the rigid end-effector link, since we know these regions cannot be occupied by objects. BHM uses a kernel method to calculate the influence of these data points to a grid of hinge points (3D points fixed in the world) by optimizing for a set of parameters $\bm{w}$. Then these sets of parameters and hinge points are used to calculate the posterior occupancy distribution.

The occupancy map can be used to query the likelihood of a point being occupied. For example the probability of a point $\bm{x}$ being occupied $y=1$ is \cite{senanayake2017bayesian}
\begin{equation}\label{eq:occupancy-prob}
\begin{split}
    P(y=1\mid x,w) &= 1-(1+\exp(\bm{w}^T\phi(\bm(x)))^{-1} \\ :&=1-\sigma(-\bm{w}^T\phi(\bm{x}))
\end{split}
\end{equation}
where $\bm{x}$ is the point being queried and $\phi(\cdot)$ is the feature vector defined as $\phi(\bm(x))=(1,k(x,\Tilde{x}_1),k(x,\Tilde{x}_2),...)$ computed between the query point and all the hinge points, and $\sigma$ is the Sigmoid function.

\begin{figure}[htb!]
  \centering
  \includegraphics[width=0.5\linewidth]{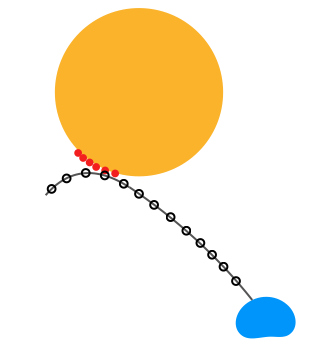}
  \caption{Location sampled along whisker shape to determine initial contact location.}
  \label{fig:sampling-prior}
\end{figure}

In order to use an occupancy map to find a localization prior, we first define a set of query points $\bm{X}_q=(\bm{x}_{q,1},\bm{x}_{q,2},...)$ which are evenly distributed along the whisker wire as illustrated in \Cref{fig:sampling-prior}. At the moment first contact is detected and the contact localization algorithm is being initialized, we can query the occupancy map at each of the points in $\bm{X}_q$ and choose the one with the highest likelihood of being occupied. Formally, this is defined as 
\begin{equation}
    \mu_0 = \arg \max_{\bm{x}}{((P(y=1\mid \bm{x}_{q,1},w),P(y=1\mid \bm{x}_{q,2},w),...)))}
\end{equation}
The design rationale for this algorithm is that, given that we assume a unique contact point, the query point with the highest likelihood of being occupied should be the most likely initial state.

\begin{figure}[htb!]
  \centering
  \includegraphics[width=\linewidth]{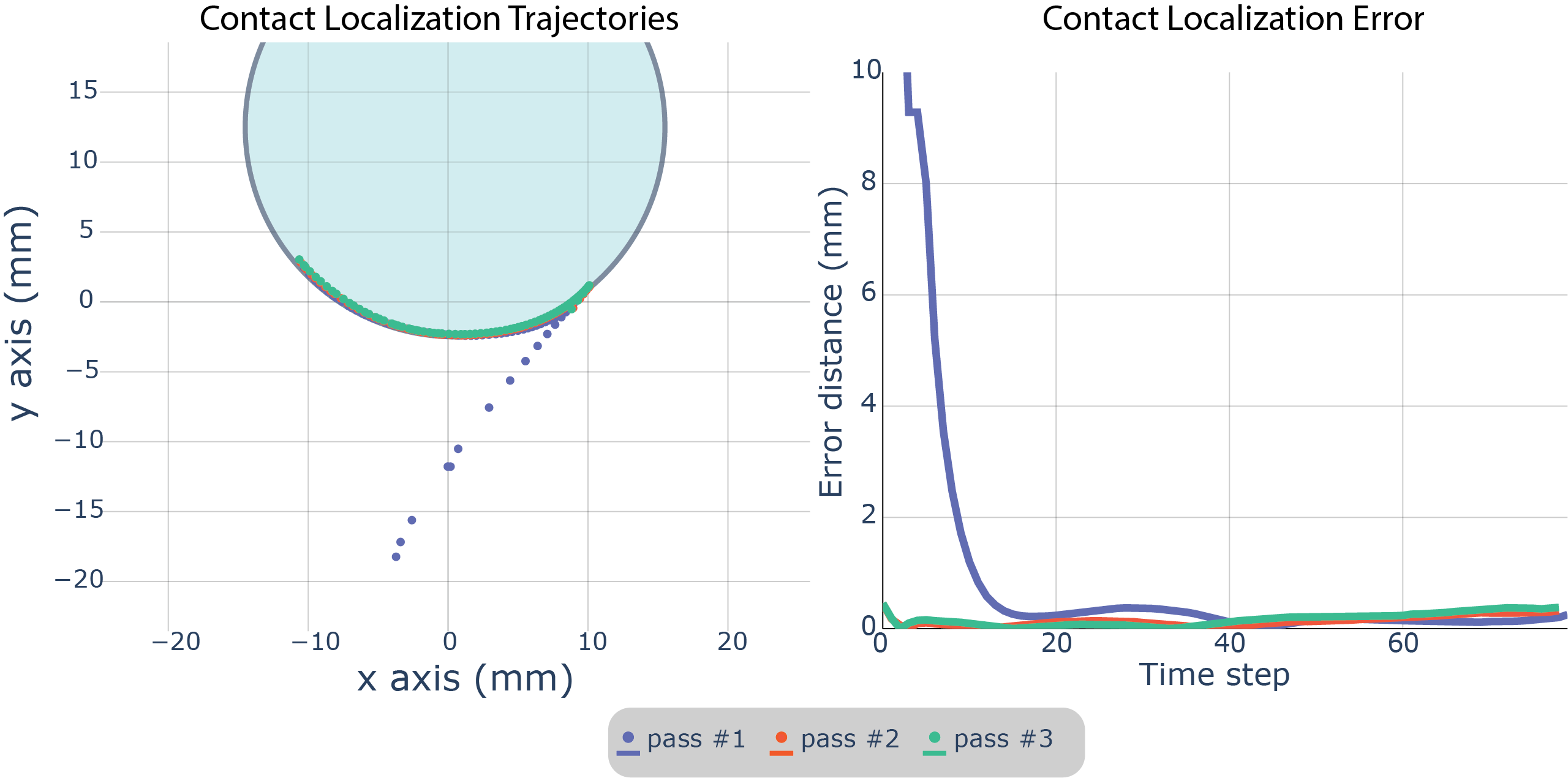}
  \caption[Contact localization convergence]{Left plot shows the contact localization result for three passes of the whisker sensors on a cylindrical object. The first pass is far from the ground truth, but the second and third pass are initialized with an occupancy map that uses prior data. Right plot shows the error over time for each of these passes, and we can appreciate that using the occupancy map can significantly reduce the error in the prior distribution.}
  \label{fig:prior}
\end{figure}

We tested this algorithm in an experiment were a whisker sensor brushed past a 30\,mm diameter cylinder multiple times. On the first pass, we make a initial guess that is approximately 2\,cm away from the ground truth, as can be seen in the blue plots in \Cref{fig:prior}. On each subsequent pass, we use the occupancy map generated using the previous pass to query for the most likely initial state. It can be seen from the plots that the second and third pass have a much smaller initial error and lead to immediate convergence to the ground truth. This experiment shows that using occupancy maps to initialize the localization algorithm can significantly improve results.

\subsection{Contact localization experiments on known shapes}
We evaluate the accuracy of our 3D contact localization method through experiments where a sensor is used to brush past an object at different heights scanning the shape of the surface. We perform this experiment on four objects of arbitrary shapes that are 3D printed with a FormLabs Form 3 printer at high resolution and positioned with an optical table relative to the sensor such that we know the shape and location of the object surface. A semi-curved sensor (i.e., with an initial straight section and curved tip, as in \Cref{fig:sampling-prior}) is used for scanning and the raw data is processed using UKF and a Thin-Plate sensor model, as described in the previous sections. 

\begin{figure}[htb!]
  \centering
  \includegraphics[width=0.85\linewidth]{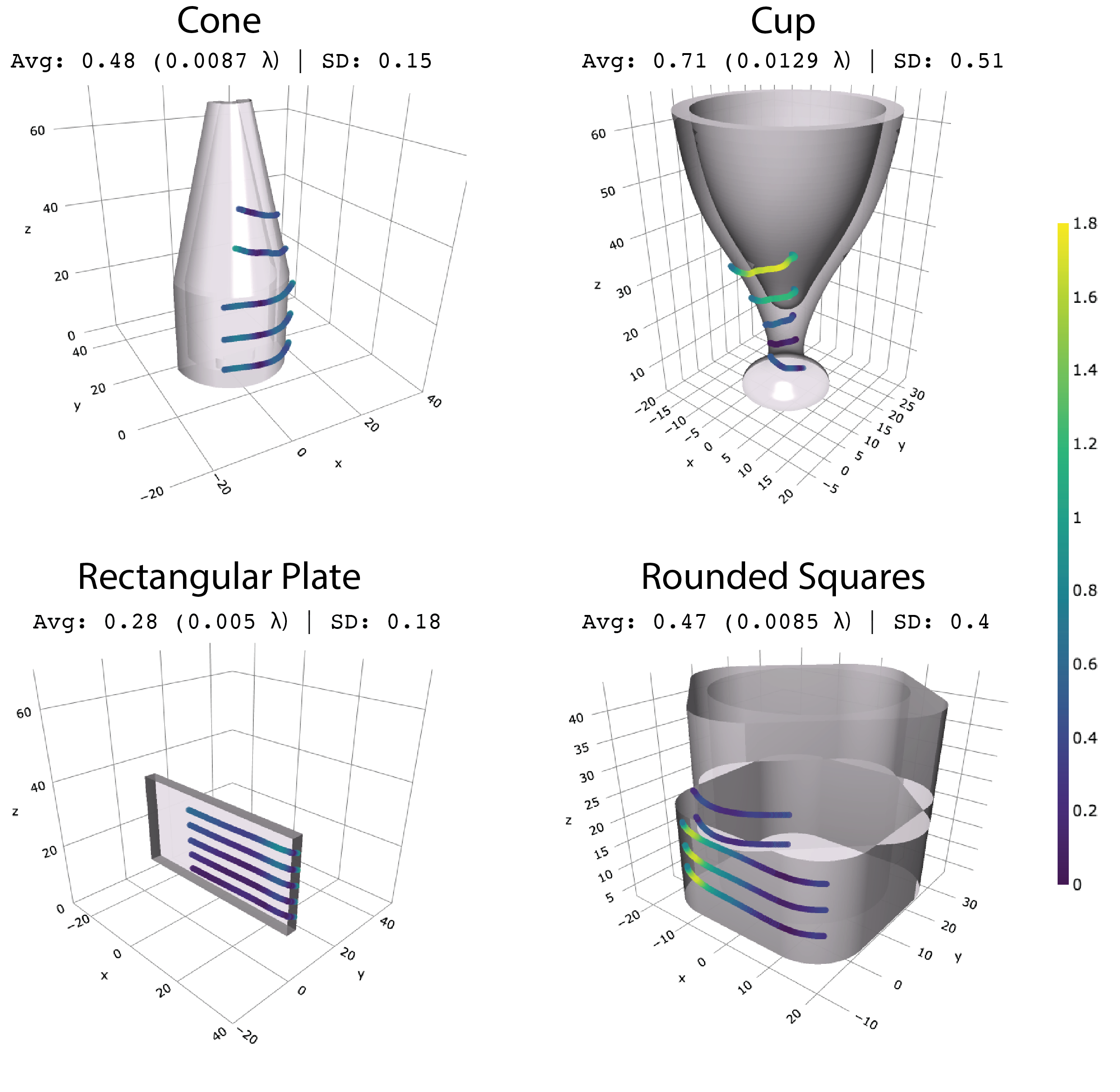}
  \caption{Contact localization on ground truth objects with known shape and location. Overlaid scatter plot shows the estimated contact location. Color and color bar shows the distance for each estimated location from the closest mesh vertex. Results for Cone object: average 0.48\,mm (S.D. 0.15), for the Cup object: average 0.71\,mm (S.D. 0.51), for the Rectangular Plate: average 0.28\,mm (S.D. 0.18), and for the Rounded Squares: average 0.47\,mm (S.D. 0.4).}
  \label{fig:object-ground-truth}
\end{figure}

Results are shown in \Cref{fig:object-ground-truth} where the 3D contact trajectories are plotted together with object meshes (fine resolution) in the world reference frame. We emphasize that there is no manual alignment of trajectories to meshes, but each of them is plotted in their absolute position. It can be seen from all the plots that the scanned data matches well with the object surface. We measured the error for each estimated contact location as its distance from the closest mesh vertex. This error for the Cone object was on average 0.48\,mm (S.D. 0.15). As a scaling factor, we can define $\lambda$ as the length of a whisker (55\,mm in the present case).  Thus the normalized error is $0.009 \lambda$. For the Cup object it was on average 0.71\,mm (S.D. 0.51) or $0.013 \lambda$. For the Rectangular Plate the error was on average 0.28\,mm (S.D. 0.18) or $0.005 \lambda$. For the Rounded Squares the error was on average 0.47\,mm (S.D. 0.4) or $0.009 \lambda$. Note that for the Cone and Cup objects it is evident that the whisker is deflected out of its principal plane of curvature, causing the contact to move in the z direction. This highlights the importance of developing a 3D localization method rather than 2D, even for motions that are predominantly in a single plane.

\begin{figure}[htb!]
  \centering
  \includegraphics[width=0.85\linewidth]{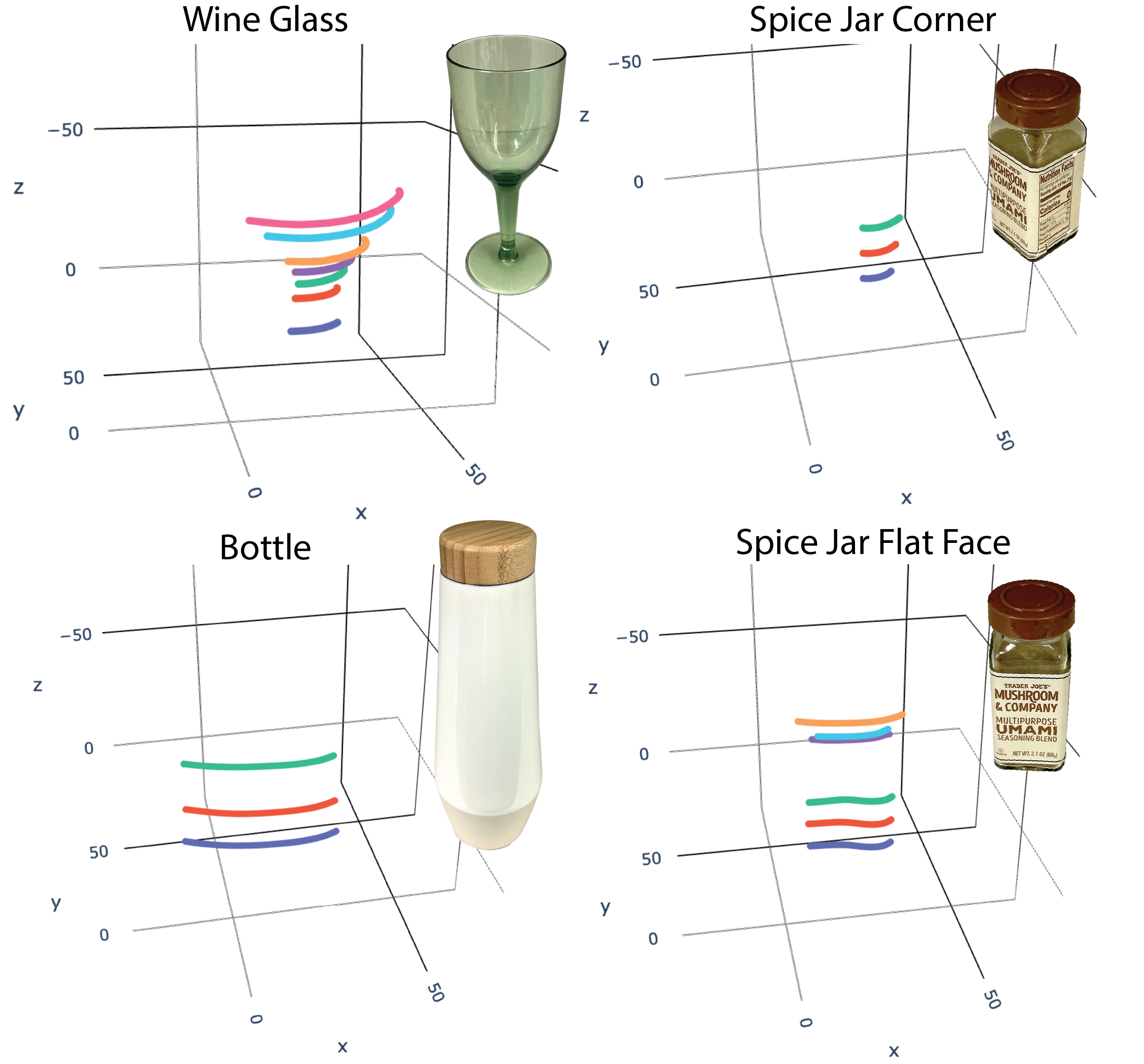}
  \caption{Contact localization on arbitrary daily objects.}
  \label{fig:object-wild}
\end{figure}

We ran the same object scanning experiment with four arbitrary objects that can typically be found in a home setting. Scanning results are shown in \Cref{fig:object-wild}. While we do not have information about the exact shape and location of these objects in the world, it can be observed from the 3D plots that the traces resemble the corresponding shapes of the object surfaces.

\subsection{Localization agnostic to whisker intrinsic shape}
An advantage of our localization method and formulation is that it is relatively agnostic to the intrinsic undeflected whisker shape. As long as we can get calibration data to produce a sensor model, it is possible to run a Bayesian Filter. This is in contrast to prior methods that need an explicit model that assumes knowledge of the shape of the whisker \cite{solomon2010extracting,nguyen2020contact,zhang2022small}. The advantage of our method is that it allows flexibility of the method to be tailored to different applications depending on sensing coverage needs. Also, our method is less susceptible to systematic errors introduced during fabrication since we perform a calibration step. To validate that our algorithm performs well for different sensor designs, we performed a simple experiment where sensors of different shapes (straight, curved and semi-curved) brush passed a single contact point and we evaluate how well can localization converge to the ground truth. The initial state estimate is purposefully placed 1\,cm away from the ground truth. Data from 10 trials was collected for each sensor design. 

\begin{figure}[htb!]
  \centering
  \includegraphics[width=0.95\linewidth]{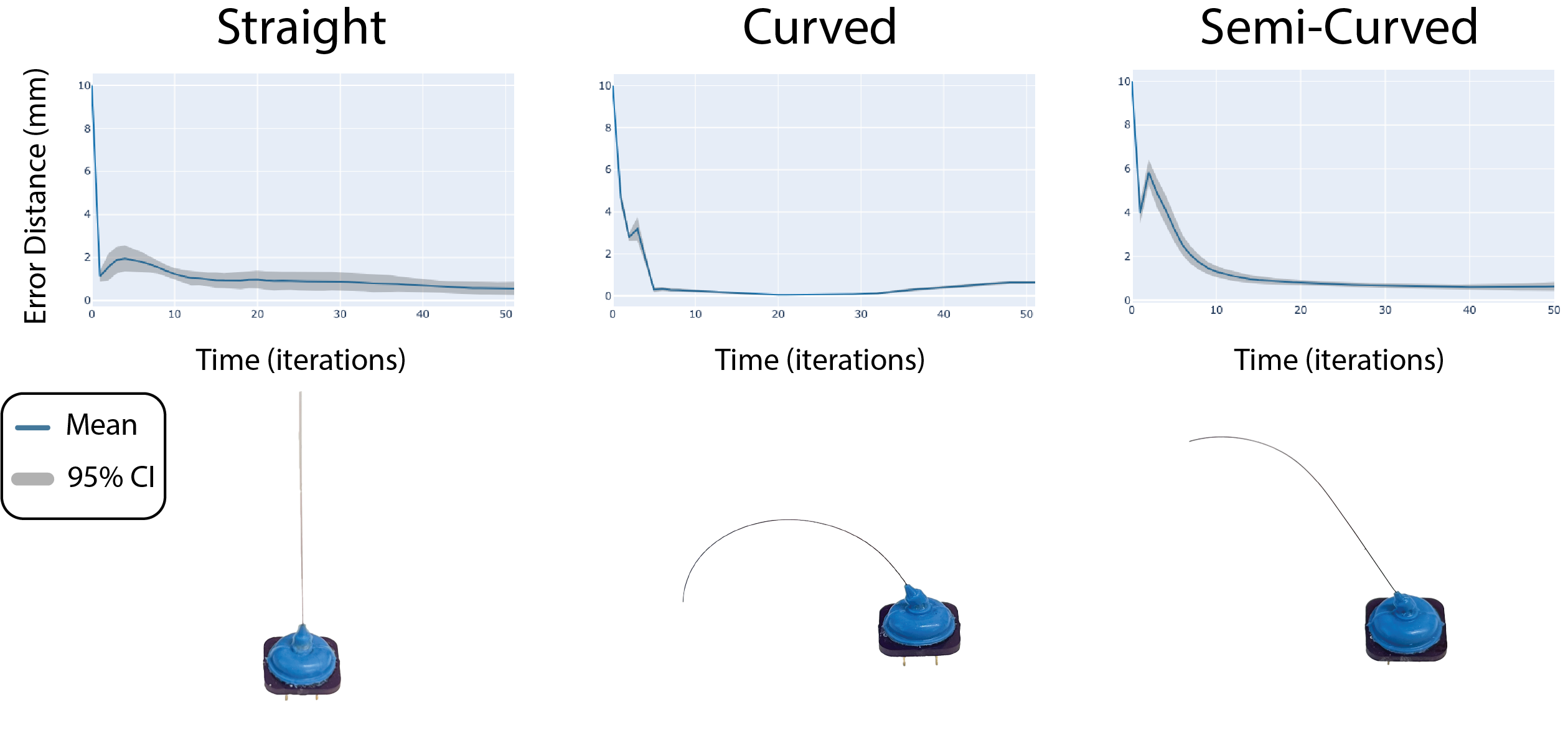}
  \caption{Contact localization performance on three sensor geometries. The experiment consisted on localizing a known contact position given an initial estimate that is off by 1 cm. Plots show distance of estimate to ground truth position for 10 trials on each sensor design case.}
  \label{fig:shape-generalization}
\end{figure}

From the results shown in \Cref{fig:shape-generalization} we can observe that tracking converges to within 1\,mm of the ground truth location in all cases. Convergence happens at different rates which might be due to the different sensor models which produce different cross-correlation matrices. This affects how large a correction step is taken in the update step as detailed in \Cref{sec:bayesian}.

\section{Mapping experiments}
\label{sec:mapping}
\subsection{Experiment setup}
In this section we demonstrate the integration of arrays of whisker sensors with an industrial robot arm in a task of reaching among objects of unknown location and shape while using our sensors and algorithms to continuously localize contacts to map the scene. The sensor system consists of 16 semi-curved sensors arranged as shown in \Cref{fig:occupancy-map}A, where each sensor is individually calibrated. The scene setup consists of a table with a collection of typical kitchen objects arranged arbitrarily. Among these objects, there are wine glasses, spice bottles, an empty vanilla extract bottle (60\,grams), and coffee mugs.

The robot arm is controlled through a joystick to move among the clutter of objects such that the whiskers on both sides brush against the objects' surfaces. During this execution we record the pose of the end-effector and the sensor readings at a rate of 250\,Hz. The data is then post-processed offline.

We used UKF to perform the contact localization with the Thin-Plate model for each sensor. The localization algorithm starts running each time contact is detected from a state of no-contact. After a sensor loses contact, the Bayesian Filter initialization parameters are reset.

\subsection{Occupancy map from contact data}
\begin{figure}
  \centering
  \includegraphics[width=0.95\linewidth]{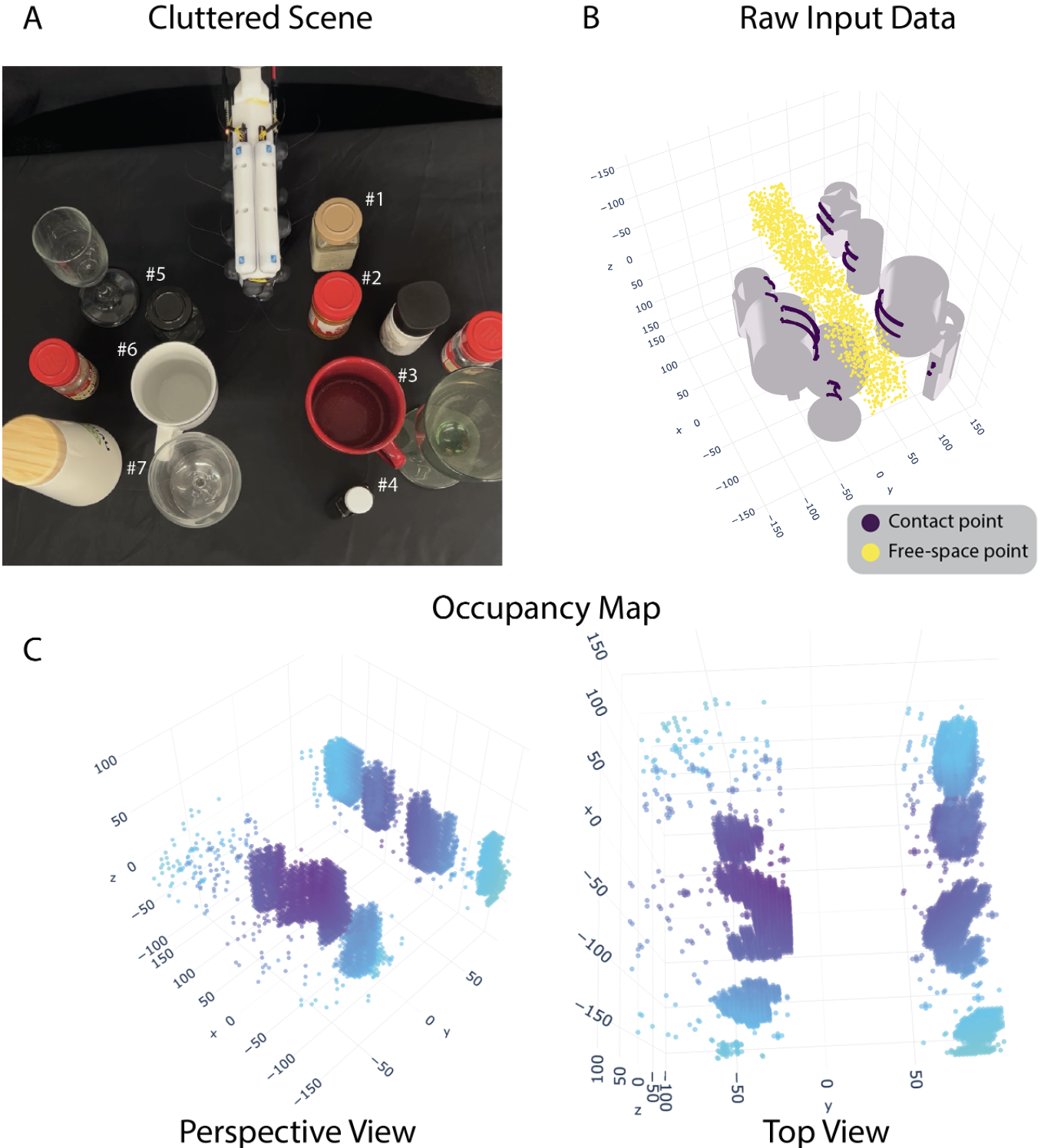}
  \caption{Bayesian Hilbert Map to reconstruct surrounding surfaces when rummaging in clutter}
  \label{fig:occupancy-map}
\end{figure}

Our sensors can accurately trace the surface of objects as shown in \Cref{fig:occupancy-map}B; however, these traces can still be too sparse to form a good picture of what the environment looks like. To interpolate this data, we use Occupancy Maps (OM) which were introduced in \Cref{sec:reusing-contacts}. Since we expect the environments the robot rummages in to present a degree of regularity -- surfaces have low spatial resolution relative to the end-effector size and objects have symmetry in shape -- it is reasonable to fill in the spaces in between curves through interpolation. Most free-standing objects in our homes such as spice jars or cans are nearly flat in the vertical direction, so this prior can be used to better interpolate contact data.

Like Gaussian Process Regression, Bayesian Hilbert Maps (BHM) also use the kernel method for related data points. Using the Squared Exponential kernel
\begin{equation}
k_{RBF}(x,\Tilde{x})=\exp{(-\gamma||x-\Tilde{x}||^2)}
\end{equation}
we can tune the smoothing parameter $\gamma$ to change the amount of relevance neighboring data points have on each other. It is also possible to choose a kernel that can incorporate prior knowledge that the length-scales of objects are larger in the vertical axis as compared to the horizontal axis (i.e., our objects' shapes tend to be vary less in the vertical direction). Such an alternative kernel is known as the Automatic Relevance Determination (ARD) RBF \cite{williams2006gaussian}, which is similar to RBF but weights the distance function with different length-scales in each dimension $\mathbf{\omega}^T(x-\tilde{x})$. We will refer to this kernel as an Elliptical Basis Function (EBF) as it is similar to a Radial Basis Function, but we use different length-scales in each dimension. The kernel is defined as follows
\begin{equation}
    k_{EBF}=(x,\Tilde{x})\exp{(-\gamma||\mathbf{\omega}^T(x-\Tilde{x})||^2)}
\end{equation}
where $\mathbf{\omega}=\begin{bmatrix}w_x\\w_y\\w_z\end{bmatrix}$ is the weight vector where each value is the reciprocal of the length-scale of its corresponding dimension. Since $x$ and $\Tilde{x}$ are defined in the world reference frame, we can choose the length scales corresponding to the vertical dimension to be higher. This would have a larger smoothing effect in the vertical direction compared to the lateral directions.

Results of the OM method are shown in \Cref{fig:occupancy-map}. \Cref{fig:occupancy-map}B shows the scatter plot of contact localization results obtained from the whisker sensors that made contact together with the unoccupied points that were randomly sampled from within the end-effector body with position estimation accuracy of 7.9\,mm. It is pertinent to note that the observed differences are primarily attributed to discrepancies between the physical placement of objects within the experimental setup and their manual delineation on the Occupancy Map. 
These two inputs are used to fit the BHM parameters $\omega$ to predict the occupancy of each point in space as shown in \Cref{eq:occupancy-prob}. To visualize the OM we sampled 3D space at intervals of 2 mm in all axes, and then plot the voxels that have a probability of being occupied of 0.9 or greater.  This result is shown in \Cref{fig:occupancy-map}C, where we can recognize all 7 objects (labeled with numbers on the photo), which are those the sensors make contact with, appearing in the reconstruction. To highlight some features that show up in the reconstruction in the Top View plot, the two coffee mugs (\#3 and \#6) are clearly shown as the two curved surfaces. The flat face of the brown spice bottle \#1 shows up as a flat surface. Sensors only make contact with the corner of spice bottle \#5 and empty plastic bottle \#4, so only a small portion of these items show up in the map. Finally, the tall wine glass \#7 is only brushed at the stem, which makes it appear as a cylinder with small radius.

\begin{figure}
  \centering
  \includegraphics[width=0.6\linewidth]{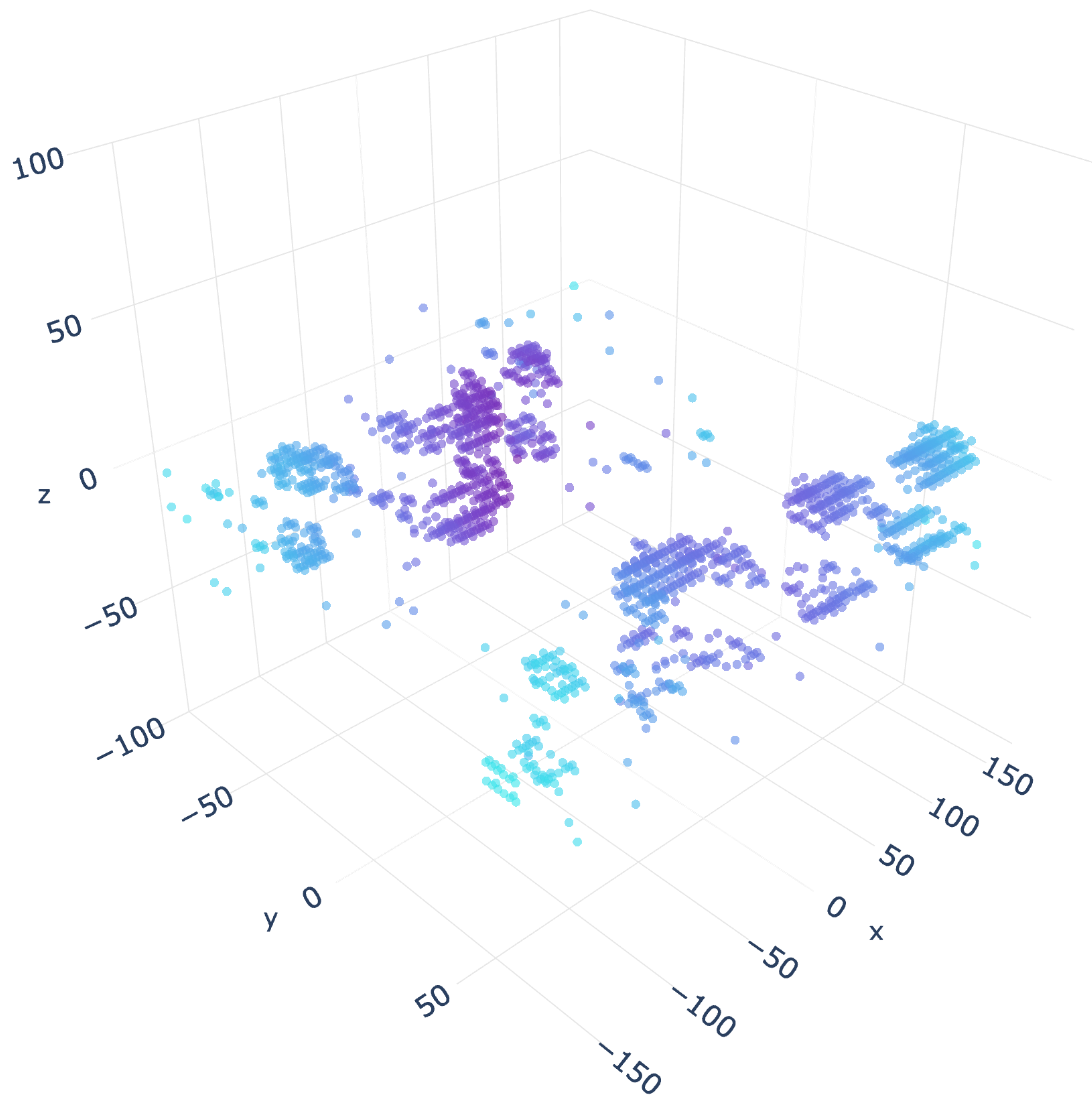}
  \caption{Surface reconstruction using Radial Basis Function (RBF), rather than Elliptical Basis Function (EBF).}
  \label{fig:occupancy-map-no-ebf}
\end{figure}

\Cref{fig:occupancy-map-no-ebf} shows an example of the surface reconstruction when using RBF instead of EBF for the OM kernel. The reconstructed surfaces only appear in two planes which is where the top and bottom whiskers on the lateral sides are located. The poor quality of the reconstruction motivates the use of the EBF kernel for mapping.

\section{Conclusion \& Future work}
\label{sec:future}
Robots reaching into unstructured environments face the challenge of unexpected contacts that may happen frequently and on any part of the robot surface. It is important to sense these contacts to help robots move safely, and to perceive information about location and shape of surrounding objects that lead to a more informed plan of robot’s motion. Tactile perception using whisker sensors is advantageous as these sensors have a low mechanical stiffness (0.17\,mNm/rad) allowing for low-force and unobtrusive contact interaction with surrounding objects. It is also possible to shape these sensors arbitrarily such that the sensing region provides a more distributed coverage of the robot link surface. We formulate a method for inferring 3D contact locations using Bayesian filtering which achieves a localization accuracy of less than 1\,mm within 0.7 seconds. This method not only improves upon prior contact localization algorithms, but also can be generalized to any curved whisker geometry. Our results show that the sensor and algorithm can enable robots to perceive local object shapes and contact locations as it navigates in proximity -- without disturbing objects. We demonstrate how it is possible to simultaneously use an array of sensors distributed on the robot surface to gather contact locations and integrate them using an Occupancy Map to reconstruct the shape of the surrounding space the robot is navigating through. In addition to perception methods, we also developed a admittance-based controller that uses real-time feedback from the sensors to avoid hard collision with nearby objects while still maintaining contact to acquire tactile data. We show how a simple controller implementation is able to create a system that easily avoids collisions with light and free-standing objects, while also being robust to object transparency, reflectivity and texture as opposed to other optical or ultrasonic proximity sensors.

Towards the future, we wish to integrate our perception and mapping method with a global motion planner in order to determine the best sequence of actions that will maximize a given objective. One such objective can be to gather information about the scene using touch, which has been explored previously using rigid tactile sensors \cite{kaboli2019tactile}. Our method has the advantage of being able to sample the space without changing its state. Using Occupancy Hilbert Map, it is possible to determine from the variance map which areas of the workspace need more tactile exploration. Another objective example is to reach a specified location in space. This can be done with traditional global path planners (e.g., graph-based planners, sampling-based planners) using the mean of the Occupancy Hilbert Map that provides the expected occupancy and delineates free-space. Balancing this two objectives is what might allow a robot to autonomously reach into, for example, a fridge and search for an object of interest with only a partial or no initial knowledge of the space. 

One interesting direction for tactile explorations with whiskers is to design the sensors to increase their sensitivity to object textures and further classify objects using this information. The current sensor design uses Nitinol wire as the whisker, which is smooth and mostly insensitive to changes in object texture. However, the design may be adapted to detect texture information as high-frequency signal content, thus not affecting contact localization.

Another direction in which our sensor and contact localization can be very useful is for mapping and localization of robots in environments with extreme visual occlusion. An example of this is underwater robots exploring a wreckage near the ocean floor under poor visibility due to muddy waters \cite{khatib2016ocean}. Having whiskers on the surface of a robot enable not only detection of imminent collisions, but also localization of contacts and tracing of the surrounding surfaces. This tactile data can be used to map the surroundings in the robot reference frame and for inferring the pose of the robot in a map. All of which may help in navigating these adverse environments.

\begin{acks}
The authors thank our collaborator Dr. Hongkai Dai as well as students Jun En Low and Rachel Thomasson for insightful discussions. We also thank Fredrik Solberg for his help with the nitinol whisker heat shaping.
\end{acks}

\begin{funding}
    This work is supported in part by TRI Global. Michael Lin was supported by the NSF GRFP Fellowship. Hao Li was supported by Zhulong Innovation Fellowship.
\end{funding}

\begin{conflict}
Mark R. Cutkosky serves on the Senior Editorial Board of the International Journal of Robotics Research. The authors declare that they have no other conflicts of interest to disclose.
\end{conflict}

\bibliographystyle{SageH}
\bibliography{citations,additional-citations}





\appendix
\begin{appendix}
\section{Appendix}
\subsection{A. Comparative Distance Sensing Details}
\label{sec:appendix}

We evaluated our whisker sensor alongside other commonly used proximity sensors, including the Vishay VCNL4010 light sensor, the STMicroelectronics VL6180X time of flight laser sensor, and the Elecfreaks HC-SR04 ultrasonic distance sensor. 

These sensors were tested on common kitchen items of differing geometry, transparency, and reflectivity, as listed in \cref{tab:compare-table}. The ``rough rock'' is obtained from an online CAD model\footnote{\url{https://www.turbosquid.com/3d-models/3d-rock-1747257 3D}} and 3D printed using Rigid 4000 Resin with a FormLab 3 printer.

All sensors were evaluated using the same apparatus depicted in \cref{fig:calibration-stage}, consisting of a multi-axis Zaber X-LRM stage with a travel distance of 100\,mm and a positioning resolution of 0.1$\mu$m.
The sensors were mounted on the stage and moved past stationary objects, at varying distances. 

During each trial, the sensor began at a position outside the object's sensing range, then moved at a speed of 1 cm/s in one direction, passing by the object to complete the trial. The minimum distance between the object and each sensor throughout the trial ranged from 2\,cm to 5\,cm, in accord with the sensing range of the whisker sensor.

Sensor signals were collected throughout each trial and invalid data were excluded in computing averages and errors. For each case we calculated the Root Mean Square (RMS) error between the measured point and the nearest point on the ground truth surface. The RMS results were averaged over all points collected along a trial. 

\end{appendix}

\end{document}